\documentclass[10pt,twocolumn,letterpaper]{article}

\usepackage[pagenumbers]{cvpr} 

\usepackage[accsupp]{axessibility}

\usepackage{microtype}
\usepackage{graphicx}
\usepackage{booktabs} 
\usepackage{multirow}
\usepackage{colortbl}
\usepackage{placeins}

\usepackage{bm}
\usepackage{nicefrac}
\usepackage{amsmath}
\usepackage{amssymb}
\usepackage{mathtools}
\usepackage{amsthm}
\usepackage{soul}
\usepackage{chngcntr}

\usepackage{booktabs}
\usepackage{multirow}
\usepackage[normalem]{ulem}
\useunder{\uline}{\ul}{}

\definecolor{mydarkorange}{HTML}{B86046}
\definecolor{codegreen}{rgb}{0,0.6,0}
\definecolor{codegray}{rgb}{0.5,0.5,0.5}
\definecolor{codepurple}{rgb}{0.58,0,0.82}
\usepackage[pagebackref=true,breaklinks=true,colorlinks,citecolor=codegreen]{hyperref}

\usepackage[utf8]{inputenc} 
\usepackage[T1]{fontenc}    
\usepackage{url}            
\usepackage{booktabs}       
\usepackage{amsfonts}       
\usepackage{nicefrac}       
\usepackage{microtype}      

\usepackage{pifont}

\usepackage[ruled,linesnumbered]{algorithm2e}
\SetKwComment{Comment}{/* }{ */}

\theoremstyle{plain}

\theoremstyle{definition}

\newtheorem{remark}{Remark}

\usepackage[misc]{ifsym}

\usepackage{enumitem}
\usepackage{soul}

\definecolor{cvprblue}{rgb}{0.21,0.49,0.74}

\def\paperID{} 
\def\confName{CVPR}
\def\confYear{2026}

\title{Where MLLMs Attend and What They Rely On: Explaining Autoregressive Token Generation}

\author{
\textbf{Ruoyu Chen$^{1,2}$, Xiaoqing Guo$^{3}$, Kangwei Liu$^{1,2}$, Siyuan Liang$^{4}$, Shiming Liu$^{5}$,}\\
~\textbf{Qunli Zhang$^{5}$, Laiyuan Wang$^{6}$, Hua Zhang$^{1,2,\textrm{\Letter}}$, Xiaochun Cao$^{7,\textrm{\Letter}}$}   \\
\small$^{1}$Institute of Information Engineering, Chinese Academy of Sciences~~\small$^{2}$University of Chinese Academy of Sciences\\
\small$^{3}$Department of Computer Science, Hong Kong Baptist University~~~~$^{4}$College of Computing and Data Science, NTU~~~~$^{5}$Huawei\\
\small$^{6}$School of Flexible Electronics, SYSU~~~~$^{7}$School of Cyber Science and Technology, Shenzhen Campus of Sun Yat-sen University\\
\small\texttt{chenruoyu@iie.ac.cn}~~~~\texttt{xiaoqingguo@hkbu.edu.hk}~~~~\texttt{liukangwei@iie.ac.cn}~~~~\texttt{pandaliang521@gmail.com}\\
\small\texttt{\{liushiming3,zhangqunli1\}@huawei.com}~~~~~~~~\texttt{zhanghua@iie.ac.cn}~~~~~~~~\texttt{caoxiaochun@mail.sysu.edu.cn}\\
Project Page: \url{https://ruoyuchen10.github.io/EAGLE/}
}

\begin{document}

\twocolumn[{%
    \maketitle
    \begin{center}
        \scriptsize
        \includegraphics[width=\textwidth]{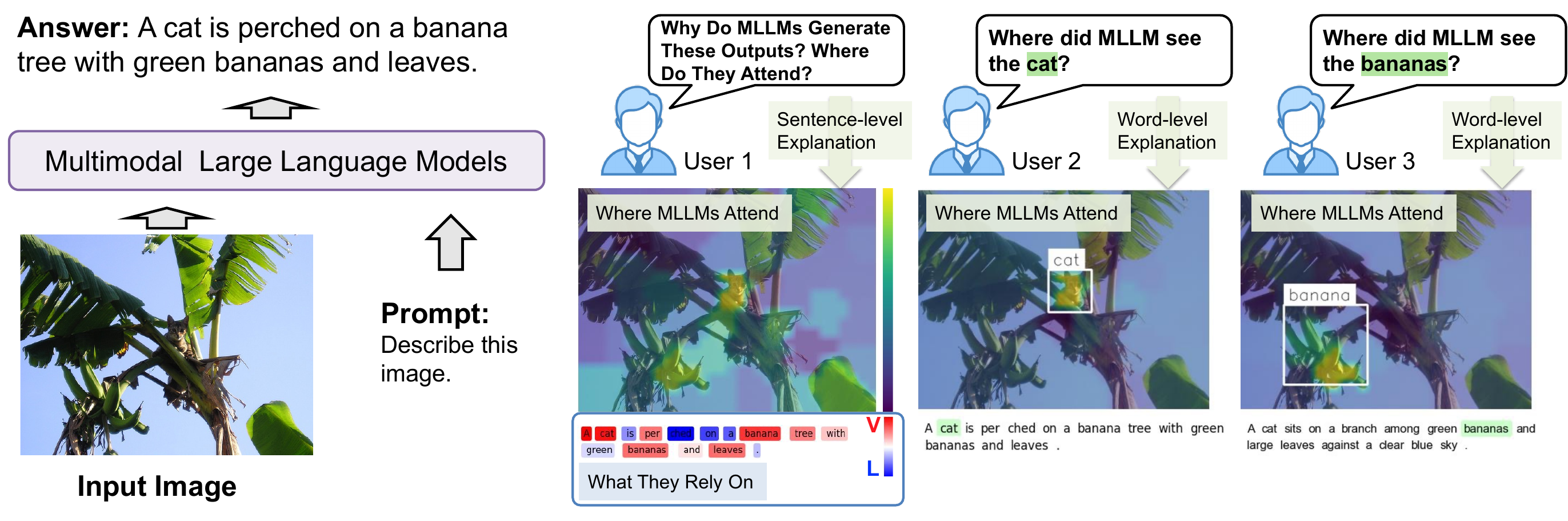}
        \vspace{-10pt}
        \captionof{figure}{\textsc{Eagle} attribution which perceptual regions drive the generation (Where MLLMs Attend) and quantifies modality reliance (What They Rely On).}
        \label{EGALE:motivation}
    \end{center}
    \vspace{2mm}
}]

\maketitle
\begin{abstract}
Multimodal large language models (MLLMs) have demonstrated remarkable capabilities in aligning visual inputs with natural language outputs. Yet, the extent to which generated tokens depend on visual modalities remains poorly understood, limiting interpretability and reliability. In this work, we present \textsc{Eagle}, a lightweight black-box framework for explaining autoregressive token generation in MLLMs. \textsc{Eagle} attributes any selected tokens to compact perceptual regions while quantifying the relative influence of language priors and perceptual evidence. The framework introduces an objective function that unifies sufficiency (insight score) and indispensability (necessity score), optimized via greedy search over sparsified image regions for faithful and efficient attribution. Beyond spatial attribution, \textsc{Eagle} performs modality-aware analysis that disentangles what tokens rely on, providing fine-grained interpretability of model decisions. Extensive experiments across open-source MLLMs show that \textsc{Eagle} consistently outperforms existing methods in faithfulness, localization, and hallucination diagnosis, while requiring substantially less GPU memory. These results highlight its effectiveness and practicality for advancing the interpretability of MLLMs. 

\end{abstract}

\renewcommand{\thefootnote}{}
\footnotetext{$\textrm{\Letter}$ Corresponding authors.}
\renewcommand{\thefootnote}{\arabic{footnote}}

\section{Introduction}
\label{sec:intro}

Multimodal large language models (MLLMs)~\citep{achiam2023gpt,wang2025internvl3,bai2025qwen2,comanici2025gemini} have achieved significant progress in vision–language understanding and generation. By jointly modeling visual and textual modalities, they can now perform a wide range of tasks, such as image captioning and visual question answering (VQA)~\citep{li2025visual}. These advances have enabled MLLMs to approach human-level performance on many benchmarks and to underpin various real-world applications~\citep{liang2024comprehensive,li2024seed,qin2026disentangle,qin2025camedit}.
However, alongside these advances come critical challenges in transparency and reliability~\citep{zhang2025redundancy}. As parameter scales and modality coverage continue to expand, MLLMs become increasingly opaque, making it difficult to trace how specific inputs influence generated outputs~\citep{xing2025large,chen2025less,chen2025interpreting}. Furthermore, MLLMs are susceptible to hallucinations~\citep{chen2025interpreting,chen2025seeing}, which undermine trust in safety-critical domains such as healthcare~\citep{ahmed2025leveraging} and autonomous driving~\citep{chen2024end}. These limitations highlight the urgent need for efficient and faithful attribution methods to improve decision transparency, diagnose errors, and enhance the safety and trustworthiness of MLLMs~\citep{lin2025survey,dang2024explainable,liang2025revisiting,liang2023badclip,xu2025defacto,liang2025safemobile,lu2025adversarial}.

Attribution in MLLMs is particularly challenging because they generate tokens autoregressively, making classification-based attribution methods difficult to adapt. Attention visualization approaches~\citep{ben2024lvlm} often fail to capture complex cross-modal interactions, while gradient-based extensions~\citep{zhang2025redundancy,xing2025large} aggregate token logits but remain confounded by textual priors. More recently, TAM~\citep{li2025token} employed activation maps to explain individual tokens and showed promising localization on Qwen2-VL~\citep{wang2024qwen2}, yet it cannot generalize to all MLLMs or capture multi-token contributions. In summary, attribution methods based on activation maps or gradients face inherent limitations: (1) activation-based approaches lack a direct causal link between inputs and outputs, reflecting only intermediate layer preferences often misaligned with human intuition; and (2) gradient-based approaches are sensitive to cumulative effects in long sequences and easily disturbed by noise and modality imbalance. The subset-selection–based VPS algorithm~\cite{chen2025interpreting} has achieved notable advances in interpreting visual grounding, outperforming gradient- and activation-based methods. However, its objective function cannot be directly transferred to MLLMs.

To more faithfully explain the generation of MLLMs, we propose \textsc{Eagle} (\textbf{E}xplaining \textbf{A}utoregressive \textbf{G}eneration by \textbf{L}anguage priors or \textbf{E}vidence), a black-box attribution framework for interpreting autoregressive token generation. As shown in Fig.~\ref{EGALE:motivation}, our method supports attribution for any chosen set of output tokens, revealing the perceptual regions that drive their generation and quantifying the relative roles of language priors and visual evidence. Inspired by VPS~\cite{chen2025interpreting}, a submodular subset selection based attribution method, we aim to find the minimal set of perceptual regions that maximizes token logits, conditioned on the prompt and context. We design an objective function with two components special for MLLMs: the insight score, capturing regions sufficient for generation, and the necessity score, identifying regions whose removal impairs generation. By applying greedy search over sparsified image regions, we construct an ordered ranking that attributes which perceptual regions promote generation in MLLMs, addressing the question of “\textbf{Where MLLMs Attend}”. Beyond spatial attribution, we also assess “\textbf{What They Rely On}”. By tracking how token logits evolve as salient regions are progressively introduced, we measure whether each token depends more on perceptual evidence or language priors, offering a faithful and comprehensive view of model decisions. 

We evaluate our method on open-source MLLMs, including LLaVA-1.5~\citep{liu2024improved}, Qwen2.5-VL~\citep{bai2025qwen2}, and InternVL3.5~\citep{wang2025internvl3}, using the MS COCO~\cite{lin2014microsoft} and MMVP~\citep{tong2024eyes} datasets for image captioning and VQA. On faithfulness metrics, our approach outperforms existing attribution methods (LLaVA-CAM~\citep{zhang2025redundancy}, IGOS++~\citep{xing2025large}, and TAM~\citep{li2025token}) by an average of 20.0\% in insertion and 13.4\% in deletion for image captioning, and by 20.6\% and 8.1\% on the same metrics for VQA. At the word level, our method achieves more rational explanations of object tokens, surpassing TAM by 36.42\% and 42.63\% on the Pointing Game under box-level and mask-level annotations, respectively. Finally, on the RePOPE benchmark~\cite{neuhaus2025repope} for object hallucination, our method accurately localizes the visual elements responsible for hallucinations and mitigates them by removing only a minimal set of interfering regions. These results demonstrate the versatility of our method across diverse tasks and benchmarks.

In summary, the contributions of this paper are:
\begin{enumerate}
    \item We propose \textsc{Eagle}, a lightweight black-box attribution framework for autoregressive token generation, which attributes any selected set of tokens to compact perceptual regions with low GPU memory cost. It further reveals the latent potential of attribution performance in MLLMs.
    \item An objective function that unifies sufficiency (insight score) and indispensability (necessity score), optimized via a greedy search strategy that balances interpretability with efficiency, yielding faithful attributions.
    \item A modality analysis that quantifies whether each generated token is driven more by language priors or perceptual evidence, enabling finer-grained interpretability.
    \item Experiments across diverse MLLMs show state-of-the-art interpretability in faithfulness, localization, and hallucination diagnosis.
\end{enumerate}

\section{Related Work}

\begin{figure*}
    \centering
    \vspace{-20pt}
    \includegraphics[width=\textwidth]{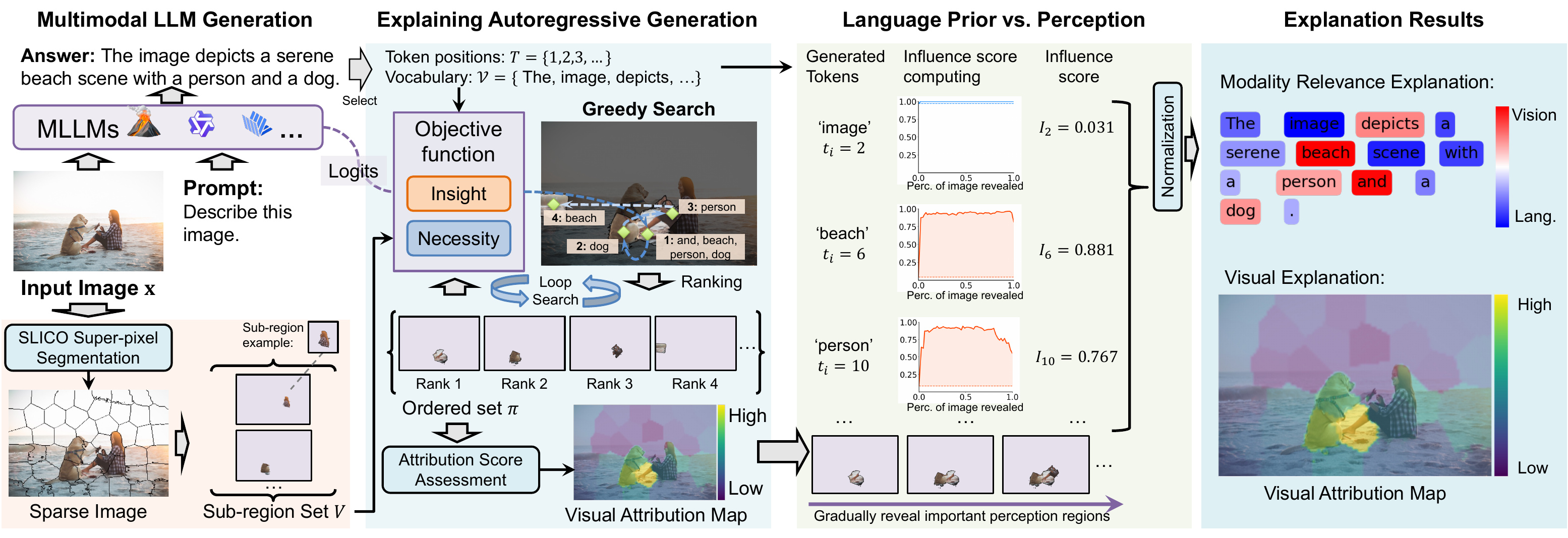}\vspace{-8pt}
    \caption{Overview of the proposed \textsc{Eagle} framework. The input image is first sparsified into sub-regions, then attributed via greedy search with the designed objective, and finally analyzed for modality relevance between language priors and perceptual evidence.} 
    \label{EGALE:framework}
    \vspace{-6pt}
\end{figure*}

\textbf{Multimodal LLMs Attribution.} Research on input-level attribution for Multimodal Large Language Models (MLLMs) is still nascent. LVLM-Interpret~\citep{ben2024lvlm} visualizes alignment between LLaVA outputs and images using raw attention, while LLaVA-CAM~\citep{zhang2025redundancy} adapts Smooth-CAM~\citep{omeiza2019smooth} to token-level probabilities, but both suffer from layer sensitivity and limited faithfulness. VPS~\citep{chen2025interpreting} introduces a search-based method for object-level tasks, yet it is restricted to grounding and detection. IGOS++\citep{xing2025large} identifies visually aligned tokens but remains parameter-sensitive. More recently, TAM\citep{li2025token} reduces contextual noise in activation maps, improving token-level attribution. However, gradient-based methods remain memory-intensive and unstable. In contrast, we propose a black-box attribution framework that localizes outputs to compact input regions without relying on token selection, quantifies the influence of language priors versus perceptual evidence, and further explains the causes of object hallucinations in MLLMs.



\textbf{Interpreting Hallucinations in MLLMs.} Several studies have applied interpretability techniques to examine hallucinations. \cite{jiang2025interpreting} investigated how image latent representations in vision-language models are projected into the language vocabulary, thereby shaping the model’s confidence in both “real” and “hallucinatory” objects, and further proposed a representation correction method to mitigate hallucinations. \cite{zhang2025mllms} examined whether MLLMs attend to incorrect regions when producing wrong answers, leveraging their internal attention maps. VaLSe~\citep{chen2025seeing} employs gradient- and attention-based attribution maps to identify noisy regions that contribute to hallucinations. In this work, we primarily focus on interpreting which input regions lead to incorrect decisions, aiming to suppress hallucinations by removing as few regions as possible.

\section{Method}

\subsection{Task Formulation}

For a multimodal large language model (MLLM), such as a VLLM, given an input image $\mathbf{x}$ and a textual prompt, the model generates an output sequence $\mathbf{y} = [y_1, y_2, \dots, y_l]$. Let $p(\cdot)$ denote the conditional probability distribution over the token vocabulary. The probability of generating each token is expressed as $p\!\left(y_t \mid \mathbf{x}, \texttt{Prompt}, \mathbf{y}_{<t}\right)$, where $\mathbf{y}_{<t} = [y_1, \dots, y_{t-1}]$ denotes the previously generated tokens. 

For interpretability analysis, inspired by VPS~\cite{chen2025interpreting}, our objective is to identify the image regions $\mathbf{x}$ that most strongly drive the model’s decisions. Image features in MLLMs are typically high-dimensional and information-dense but also redundant and less directly interpretable than text. We therefore focus on decomposing $\mathbf{x}$ into semantically meaningful subregions. Specifically, the image is sparsified into $V = \{\mathbf{x}_{1}, \mathbf{x}_{2}, \dots, \mathbf{x}_{N}\}$ using the SLICO~\citep{achanta2012slic} superpixel segmentation method, where $\mathbf{x}_{i}$ denotes the $i$-th subregion. The attribution problem is then cast as a subset selection task~\citep{chen2024less}:
$\max_{S \subseteq V, |S| < k} \mathcal{F}(S)$,
where $k$ is the maximum number of selected subregions and $\mathcal{F}(\cdot)$ is a set function measuring interpretability. Beyond the unordered case, attribution also depends on the order in which regions contribute to the decision. We therefore extend the formulation to ordered subsets:
\begin{equation}\label{EAGLE:object}
\max_{\pi \in \mathcal{P}(V), |\pi| < k} \sum_{r=1}^{|\pi|} \mathcal{F}(\pi_{:r}),
\end{equation}
where $\pi$ is an ordered subset, $\mathcal{P}(V)$ the collection of all ordered subsets of $V$, and $r$ the prefix length. The problem thus reduces to designing $\mathcal{F}(\cdot)$ and optimizing it efficiently.

\subsection{Explaining Autoregressive Generation}\label{EAGLE:EAG}

We propose \textsc{Eagle}, a novel attribution framework for explaining autoregressive token generation, as shown in Fig.~\ref{EGALE:framework}. For the set function in Eq.~\ref{EAGLE:object}, we design a submodular-inspired objective to measure interpretability. This objective encourages diminishing returns as more regions are added, although it may not be strictly submodular for MLLMs. Let $T = [t_1, t_2, \dots, t_n]$ denote the token positions of interest, and $\mathcal{V} = [v_1, v_2, \dots, v_n]$ their corresponding vocabulary indices.

\textbf{Insight Score:} A key metric for interpretability is the identification of the minimal set of input regions sufficient to maximize the probability of generating the target label, thereby highlighting the most informative evidence underlying the model’s decision. Given an input prompt and an image $\mathbf{x}$, we denote the corresponding target sequence as $\mathbf{y}$, which is generated conditioned on both. For a candidate subregion $S$, the insight score is defined as:
\begin{equation}\label{EAGLE:insight}
    \begin{aligned}
    s_{\text{insight}}&(S,\texttt{Prompt},\mathbf{y},T,\mathcal{V}) =\\
    &\sum_{i=1}^{n}
    p\!\left(y_{t_i}=v_i \mid S,\texttt{Prompt},\mathbf{y}_{<t_i}\right),
    \end{aligned}
\end{equation}
where $p(y_{t_i} = v_i \mid S, \texttt{Prompt}, \mathbf{y}_{<t_i})$ denotes the probability of generating the ground-truth token $y_{t_i}$ at position $t_i$, conditioned on the selected subregion $S$, the input prompt, and the previously generated tokens.

\textbf{Necessity Score:} Another key metric for interpretability is the identification of the minimal set of input regions whose removal leads to a significant decrease in the probability of generating the target label, thereby revealing the indispensable evidence that the model relies on. Formally, the necessity score is defined as:
\begin{equation}\label{EAGLE:necessity}
    \begin{aligned}
    s_{\text{necessity}}&(V \setminus S,\texttt{Prompt},\mathbf{y},T,\mathcal{V}) 
    = \\
    &\sum_{i=1}^{n} \Big( 1 - p\!\left(y_{t_i} = v_i \mid V \setminus S, \texttt{Prompt}, \mathbf{y}_{<t_i}\right) \Big),
    \end{aligned}
\end{equation}
where $V \setminus S$ denotes the remaining regions after removing $S$. This score provides an effective criterion in the search phase for uncovering subtle but critical regions that contribute to the final decision.

\textbf{Objective Function:} We integrate the insight and necessity scores into a unified objective function that jointly captures sufficiency and necessity for interpreting autoregressive token generation:
\begin{equation}\label{EAGLE:overall}
    \begin{aligned}
    \mathcal{F}(S,V,\texttt{Prompt},&\mathbf{y},T,\mathcal{V}) 
    = s_{\text{insight}}(S,\texttt{Prompt},\mathbf{y},T,\mathcal{V}) \\
    &+ s_{\text{necessity}}(V \setminus S,\texttt{Prompt},\mathbf{y},T,\mathcal{V}),
    \end{aligned}
\end{equation}
where a larger objective value indicates that the selected input combination $S$ is more important and thus provides stronger interpretability.

\textbf{Saliency Map Generation:} Similar to the calculation process for VPS~\cite{chen2025interpreting}, to optimize the objective in Eq.~\ref{EAGLE:object}, an $\mathcal{NP}$-hard problem, a greedy search strategy is adopted. Following~\cite{chen2025interpreting}, we assess saliency contrast across subregions via marginal increments of the objective evaluated along the ordered sequence. Subregions yielding larger increments are deemed influential; diminishing increments approaching zero suggest that later subregions add little information and exhibit weak saliency differentiation.

\subsection{Language Prior vs. Perception Evidence}

Beyond identifying which perceptual regions promote the generation of specific autoregressive tokens, we further analyze whether each generated token is more strongly influenced by language priors or by perceptual evidence. Existing approaches often assess token relevance to the visual modality by observing changes in probability when the input image is masked~\cite{xing2025large}. However, simply comparing the probability with the full image against that without the image is not a reliable indicator of visual relevance, as the probability may first increase and then decrease when visual inputs are progressively inserted~\citep{chen2024less}. By contrast, if a token is truly irrelevant to the visual modality, its probability should remain stable regardless of how the image is modified.

To address this limitation, we leverage the ordered subset $\pi$ obtained in Section~\ref{EAGLE:EAG} and examine how each token is affected as the subregions in $\pi$ are progressively expanded, thereby quantifying the extent to which the token is influenced by perceptual evidence. Specifically, for each target token position $t_i \in T$, the influence score is defined as:
\begin{equation}
    \begin{aligned}
    I_{t_i}
    &= \sum_{r=1}^{|\pi|} \Big(
    p\!\left(y_{t_i}=v_i \mid \pi_{:r}, \texttt{Prompt}, \mathbf{y}_{<t_i}\right)\\
    &\quad - \min_{1 \le j \le |\pi|}
    p\!\left(y_{t_i}=v_i \mid \pi_{:j}, \texttt{Prompt}, \mathbf{y}_{<t_i}\right)
    \Big).
    \end{aligned}
\end{equation}
where $v_{i}$ denotes the vocabulary index of the target token $y_{t_i}$. The influence score $I_{t_i}$ measures the impact of perceptual evidence on the generation of token $y_{t_i}$. A larger score indicates that the token generation is more strongly driven by perceptual evidence, whereas a smaller score suggests a greater reliance on language priors, as shown in Fig.~\ref{EGALE:framework}. The detailed calculation process of the proposed \textsc{Eagle} algorithm is outlined in the \textit{supplementary materials}


\begin{table*}[!t]
    \caption{Evaluation of sentence-level faithfulness metrics (Deletion, Insertion AUC, and Average Highest Score) on the MS COCO and MMVP datasets using LLaVA-1.5, Qwen2.5-VL, and InternVL3.5.}\vspace{-12pt}
    \label{EAGLE:faithfulness_on_sentence_level}
    \begin{center}
        \resizebox{\textwidth}{!}{
            \begin{tabular}{lcl|ccc|ccc|cc}
            \toprule[1.5pt]
            \multirow{2}{*}{Datasets} & \multirow{2}{*}{MLLMs} & \multirow{2}{*}{Methods}  & \multicolumn{3}{c|}{Sentence-level Faithfulness} & \multicolumn{3}{c|}{Sensitive Tokens-level Faithfulness} & \multirow{2}{*}{GPU Memory ($\downarrow$)} & \multirow{2}{*}{Time Consum. ($\downarrow$)} \\ 
             &  &   & Ins. ($\uparrow$) & Del. ($\downarrow$) & Ave. high. score ($\uparrow$) & Ins. ($\uparrow$)  & Del. ($\downarrow$) & Ave. high. score ($\uparrow$) \\ \midrule
            \multirow{12}{*}{\begin{tabular}[c]{@{}l@{}}MS COCO~\citep{lin2014microsoft}\\ (Image caption task)\end{tabular}} & \multirow{3}{*}{LLaVA-1.5 7B~\citep{liu2024improved}} & LLaVA-CAM~\citep{zhang2025redundancy}  & 0.5298 & 0.5317 & 0.6031 & 0.4124 & 0.4115 & 0.5783 & 37.25 GB & 15.4 s \\
            & & iGOS++ (w/ GNC)~\citep{xing2025large}  & 0.5293 & 0.5168 & 0.6004 & 0.4101 & 0.3815 & 0.5731 & 48.18 GB & 36.3 s \\
            & & \cellcolor[HTML]{EFEFEF}\textsc{Eagle} & \cellcolor[HTML]{EFEFEF}\textbf{0.5970} & \cellcolor[HTML]{EFEFEF}\textbf{0.4554} & \cellcolor[HTML]{EFEFEF}\textbf{0.6259} & \cellcolor[HTML]{EFEFEF}\textbf{0.5344} & \cellcolor[HTML]{EFEFEF}\textbf{0.2809} & \cellcolor[HTML]{EFEFEF}\textbf{0.5993} & \cellcolor[HTML]{EFEFEF}\textbf{16.07 GB} & \cellcolor[HTML]{EFEFEF}258.4 s \\
            \cmidrule(l){2-11} 
             & \multirow{3}{*}{Qwen2.5-VL 3B~\citep{bai2025qwen2}} & LLaVA-CAM~\citep{zhang2025redundancy}  & 0.4978 & 0.5562 & 0.6662 & 0.3541 & 0.4497 & 0.6424 & 28.99 GB & 22.2 s \\
            & & iGOS++ (w/ GNC)~\citep{xing2025large}  & 0.5328 & 0.4891 & 0.6672 & 0.4021 & 0.3273 & 0.6473 & 71.62 GB & 46.6 s \\
            & & \cellcolor[HTML]{EFEFEF}\textsc{Eagle} & \cellcolor[HTML]{EFEFEF}\textbf{0.6479} & \cellcolor[HTML]{EFEFEF}\textbf{0.4345} & \cellcolor[HTML]{EFEFEF}\textbf{0.7039} & \cellcolor[HTML]{EFEFEF}\textbf{0.5867} & \cellcolor[HTML]{EFEFEF}\textbf{0.2710} & \cellcolor[HTML]{EFEFEF}\textbf{0.6840} & \cellcolor[HTML]{EFEFEF}\textbf{8.75 GB} & \cellcolor[HTML]{EFEFEF}418.3 s \\
            \cmidrule(l){2-11} 
            & \multirow{3}{*}{Qwen2.5-VL 7B~\citep{bai2025qwen2}} & LLaVA-CAM~\citep{zhang2025redundancy} & 0.5605 & 0.5464 & 0.7235 & 0.4467 & 0.4209 & 0.7010 & 47.17 GB & 21.8 s \\
            & & iGOS++ (w/ GNC)~\citep{xing2025large} & 0.5603 & 0.5072 & 0.7237 & 0.4400 & 0.3623 & 0.6695 & 96.90 GB & 62.7 s \\
            & & \cellcolor[HTML]{EFEFEF}\textsc{Eagle} & \cellcolor[HTML]{EFEFEF}\textbf{0.7006} & \cellcolor[HTML]{EFEFEF}\textbf{0.4597} & \cellcolor[HTML]{EFEFEF}\textbf{0.7578} & \cellcolor[HTML]{EFEFEF}\textbf{0.6337} & \cellcolor[HTML]{EFEFEF}\textbf{0.2988} & \cellcolor[HTML]{EFEFEF}\textbf{0.7285} & \cellcolor[HTML]{EFEFEF}\textbf{17.68 GB} & \cellcolor[HTML]{EFEFEF}436.2 s\\
            \cmidrule(l){2-11} 
            & \multirow{3}{*}{InternVL3.5 4B~\citep{wang2025internvl3}} & LLaVA-CAM~\citep{zhang2025redundancy}  & 0.6116 & 0.6235 & 0.8032 & 0.4948 & 0.5100 & 0.7764 & 81.84 GB & 54.3 s \\
            & & iGOS++ (w/ GNC)~\citep{xing2025large} & 0.6271 & 0.5726 & 0.7999 & 0.5088 & 0.4337 & 0.7715 & 60.93 GB & 40.1 s \\
            & & \cellcolor[HTML]{EFEFEF}\textsc{Eagle} & \cellcolor[HTML]{EFEFEF}\textbf{0.7665}  & \cellcolor[HTML]{EFEFEF}\textbf{0.4650} &  \cellcolor[HTML]{EFEFEF}\textbf{0.8335} & \cellcolor[HTML]{EFEFEF}\textbf{0.7042} & \cellcolor[HTML]{EFEFEF}\textbf{0.3042} & \cellcolor[HTML]{EFEFEF}\textbf{0.8051} & \cellcolor[HTML]{EFEFEF}\textbf{12.45 GB} & \cellcolor[HTML]{EFEFEF}803.5 s \\
            \midrule
            \multirow{12}{*}{\begin{tabular}[c]{@{}l@{}}MMVP~\citep{tong2024eyes}\\ (VQA task)\end{tabular}} & \multirow{3}{*}{\begin{tabular}[c]{@{}c@{}}LLaVA-1.5 7B \citep{liu2024improved}\end{tabular}} & LLaVA-CAM~\citep{zhang2025redundancy} & 0.7756 & 0.7745 & 0.7980 & 0.6076 & 0.6044 & 0.7275 & 34.38 GB & 12.2 s \\
            & & iGOS++ (w/ GNC)~\citep{xing2025large} & 0.7717 & 0.7698 & 0.7965 & 0.5825 & 0.5781 & 0.7236 & 92.90 GB & 34.8 s \\
            & & \cellcolor[HTML]{EFEFEF}\textsc{Eagle} & \cellcolor[HTML]{EFEFEF}\textbf{0.7960} & \cellcolor[HTML]{EFEFEF}\textbf{0.7474} & \cellcolor[HTML]{EFEFEF}\textbf{0.8086} & \cellcolor[HTML]{EFEFEF}\textbf{0.6867} & \cellcolor[HTML]{EFEFEF}\textbf{0.5027} &  \cellcolor[HTML]{EFEFEF}\textbf{0.7507} & \cellcolor[HTML]{EFEFEF}\textbf{15.40 GB} & \cellcolor[HTML]{EFEFEF}252.6 s \\
            \cmidrule(l){2-11} 
            & \multirow{3}{*}{\begin{tabular}[c]{@{}c@{}}Qwen2.5-VL 3B \citep{bai2025qwen2}\end{tabular}} & LLaVA-CAM~\citep{zhang2025redundancy} & 0.7742 & 0.7770 & 0.8181  & 0.5925 & 0.6006 & 0.7476 & 19.17 GB & 16.1 s \\
            & & iGOS++ (w/ GNC)~\citep{xing2025large} & 0.7719 & 0.7613 & 0.8183 & 0.5719 & 0.5356 & 0.7437 &  19.79 GB & 31.8 s \\
            & & \cellcolor[HTML]{EFEFEF}\textsc{Eagle} & \cellcolor[HTML]{EFEFEF}\textbf{0.8052}       & \cellcolor[HTML]{EFEFEF}\textbf{0.7338}       & \cellcolor[HTML]{EFEFEF}\textbf{0.8339}      & \cellcolor[HTML]{EFEFEF}\textbf{0.6634}         & \cellcolor[HTML]{EFEFEF}\textbf{0.4935} &  \cellcolor[HTML]{EFEFEF}\textbf{0.7689} & \cellcolor[HTML]{EFEFEF}\textbf{8.76 GB} & \cellcolor[HTML]{EFEFEF}222.7 s \\
            \cmidrule(l){2-11} 
            & \multirow{3}{*}{\begin{tabular}[c]{@{}c@{}}Qwen2.5-VL 7B \citep{bai2025qwen2}\end{tabular}} & LLaVA-CAM~\citep{zhang2025redundancy} &       0.7505     &    0.7486    &  0.8042 & 0.4974 &      0.4847     &     0.7242 & 37.54 GB & 14.5 s \\
            & & iGOS++ (w/ GNC)~\citep{xing2025large} & 0.7394 & 0.7211 & 0.8036 & 0.4505 & 0.3853 & 0.7185 & 32.76 GB & 31.0 s  \\
            & & \cellcolor[HTML]{EFEFEF}\textsc{Eagle} &   \cellcolor[HTML]{EFEFEF}\textbf{0.7824}  & \cellcolor[HTML]{EFEFEF}\textbf{0.6996}   &   \cellcolor[HTML]{EFEFEF}\textbf{0.8119}   & \cellcolor[HTML]{EFEFEF}\textbf{0.5901}    &  \cellcolor[HTML]{EFEFEF}\textbf{0.3675}    & \cellcolor[HTML]{EFEFEF}\textbf{0.7362} & \cellcolor[HTML]{EFEFEF}\textbf{17.40 GB} & \cellcolor[HTML]{EFEFEF}220.8 s \\
            \cmidrule(l){2-11} 
            & \multirow{3}{*}{\begin{tabular}[c]{@{}c@{}}InternVL3.5 4B \citep{wang2025internvl3}\end{tabular}} & LLaVA-CAM~\citep{zhang2025redundancy} & 0.7348 & 0.7458 & 0.8325 & 0.4897 & 0.5213 & 0.7575 & 27.20 GB & 20.7 s \\
            & & iGOS++ (w/ GNC)~\citep{xing2025large} & 0.7277 & 0.7160 & 0.8302 & 0.4743 & 0.4454  & 0.7535 & 62.31 GB & 34.4 s \\
            & & \cellcolor[HTML]{EFEFEF}\textsc{Eagle} & \cellcolor[HTML]{EFEFEF}\textbf{0.8012} & \cellcolor[HTML]{EFEFEF}\textbf{0.6782} &  \cellcolor[HTML]{EFEFEF}\textbf{0.8471} & \cellcolor[HTML]{EFEFEF}\textbf{0.6379} & \cellcolor[HTML]{EFEFEF}\textbf{0.4027} & \cellcolor[HTML]{EFEFEF}\textbf{0.7762} & \cellcolor[HTML]{EFEFEF}\textbf{12.26 GB} & \cellcolor[HTML]{EFEFEF}176.5 s \\
            \bottomrule[1.5pt]
            \end{tabular}
        }
    \end{center}
    \vspace{-20pt}
\end{table*}

\begin{remark}[Token-Agnostic Attribution]
    Gradient-based methods~\citep{xing2025large} rely on selecting visually relevant tokens; choosing tokens dominated by language priors can distort attribution and yield unreliable explanations. 
    In contrast, our approach is token-agnostic: even when applied to tokens strongly influenced by language priors, the visual attribution remains unaffected. Moreover, after attribution, our framework explicitly evaluates whether the selected tokens are primarily driven by perceptual evidence or language priors.
\end{remark}

\begin{remark}[Interactive Explanation]
    Our framework also allows users to select specific sentences, words, or tokens for targeted attribution. 
    This flexibility enables fine-grained interpretation at arbitrary granularity and naturally supports human-in-the-loop analysis and interactive explanation.
\end{remark}

\begin{remark}[Computational Complexity]
The algorithm has time complexity $\mathcal{O}(2^{|V|})$. With the greedy strategy, all subregions are ordered with a total of $\tfrac{1}{2}|V|^2 + \tfrac{1}{2}|V|$ inferences, yielding a time complexity of $\mathcal{O}(|V|^2)$. The space complexity is $\mathcal{O}(|V|)$, only the ordered subset needs to be stored.
\end{remark}


\section{Experiments}

\subsection{Experimental Setup}

\textbf{Datasets.} We evaluate across three representative tasks: MS COCO Caption~\citep{lin2014microsoft,chen2015microsoft} for image captioning, MMVP~\citep{tong2024eyes} for visual question answering (VQA), and RePOPE~\citep{neuhaus2025repope} for object hallucination assessment.

\textbf{Baselines.} We compare \textsc{Eagle} against state-of-the-art attribution methods for MLLMs, including gradient-based approaches (LLaVA-CAM~\citep{zhang2025redundancy} and IGOS++ adaptation~\citep{xing2025large}) and the activation-based method TAM~\citep{li2025token}. Note that TAM is restricted to attributing a single token at a time and cannot handle token combinations.

\textbf{Models.} We validate our approach on three multimodal large language models: LLaVA-1.5-7B~\citep{liu2024improved}, Qwen2.5-VL (3B and 7B)~\citep{bai2025qwen2}, and InternVL 3.5-4B~\citep{wang2025internvl3}.

\textbf{Evaluation Metrics.} 
We consider three categories of attribution metrics: \emph{faithfulness}, \emph{localization}, and \emph{correction-oriented}. 
(1) Faithfulness metrics evaluate whether explanations align with the model’s decision process. We adopt \emph{Insertion}~\citep{petsiuk2018rise}, \emph{Deletion}~\citep{petsiuk2018rise}, and \emph{Average Highest Score}~\citep{chen2024less}, computed as the mean probability over selected tokens. 
(2) Localization metrics assess whether explanations overlap with ground-truth regions using the \emph{Point Game}~\citep{zhang2018top}, under both \emph{box-level} and \emph{mask-level} annotations, where correctness is defined by the maximum attribution point falling inside the bounding box or segmentation mask. 
(3) Correction-oriented metrics address hallucination evaluation by testing whether attributions reveal regions causing hallucinated outputs. We use \emph{Average Minimal Correction Region (AMCR)}, the average proportion of regions that must be removed to correct hallucinations, and \emph{Correction Success Rate under Budget (CSR@10\%)}, the percentage of cases corrected when no more than 10\% of regions are removed. 

\begin{figure*}
    \centering
    \vspace{-20pt}
    \includegraphics[width=\textwidth]{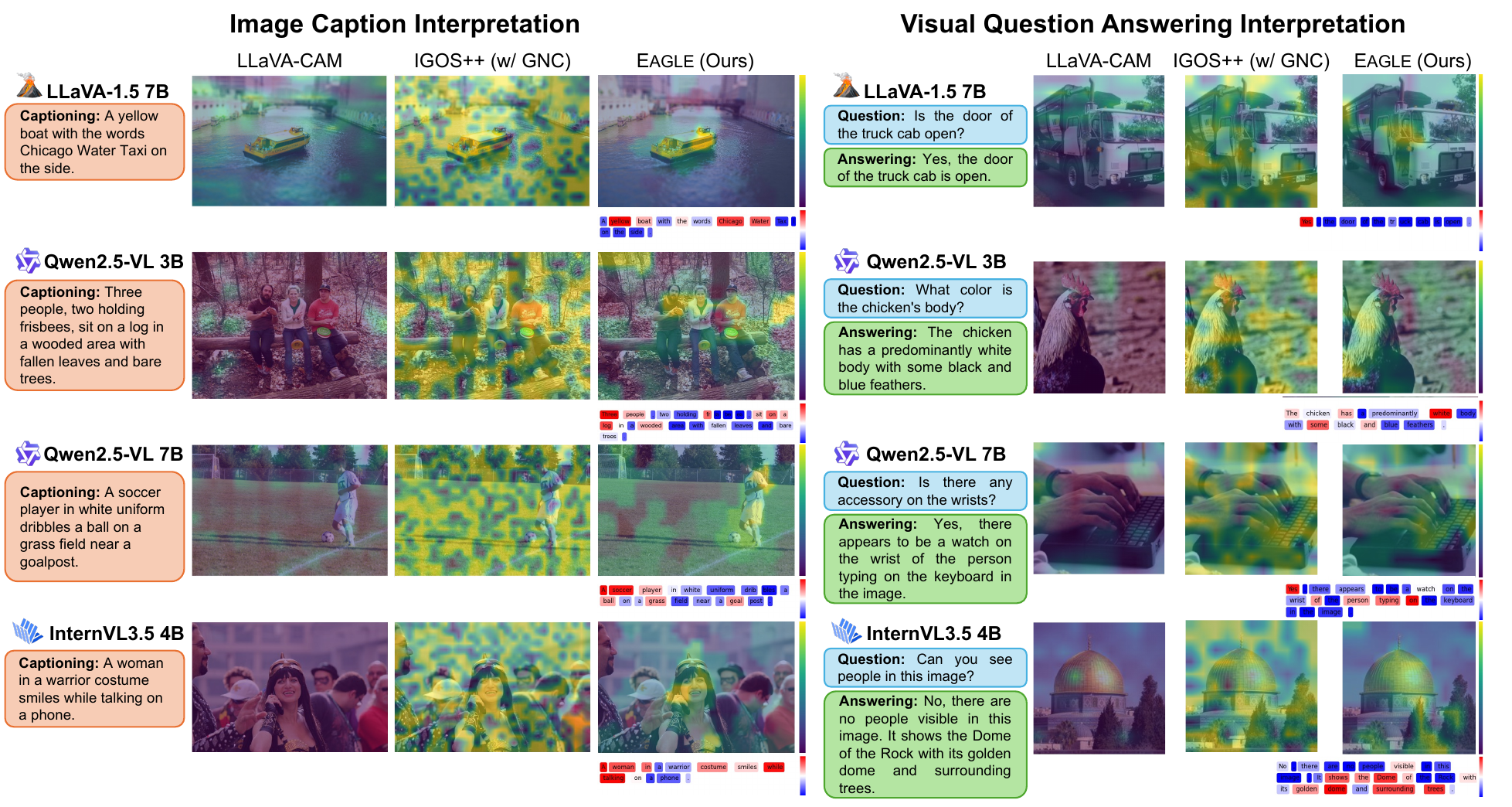}\vspace{-8pt}
    \caption{Visualization of explanation results for LLaVA-1.5, Qwen2.5-VL, and InternVL3.5 on the MS COCO and MMVP datasets.} 
    \label{EGALE:sentence-level-visualization}
    \vspace{-5pt}
\end{figure*}

\subsection{Faithfulness on Sentence-level Explanations}

We begin by evaluating our attribution method on two common MLLM tasks, image captioning and visual question answering (VQA), with the goal of identifying which image regions drive the full content generated by the model. We primarily compare our approach against LLaVA-CAM~\citep{zhang2025redundancy} and IGOS++ (w/ GNC)~\citep{xing2025large}. Table~\ref{EAGLE:faithfulness_on_sentence_level} reports results on faithfulness metrics, evaluated in two ways: (1) using the sum of logits over all predicted tokens, and (2) using the sum over sensitive tokens, defined as those whose logits change by more than 0.2 when the entire image is masked.

For the image captioning task, our method consistently achieves state-of-the-art performance across all models and metrics. On the LLaVA-1.5 7B model, it surpasses the best results of LLaVA-CAM and IGOS++ (w/ GNC) by 12.7\%, 11.9\%, and 3.8\% in sentence-level insertion, deletion, and average highest score, respectively. At the sensitive-token level, the improvements are even larger, reaching 29.6\%, 26.3\%, and 3.6\%. These stronger gains arise because sensitive tokens are more strongly grounded in visual evidence, making them particularly responsive to well-localized attribution maps. Similar trends are observed on the Qwen2.5-VL 7B model, where our method improves over the best baselines by 25.0\%, 9.4\%, and 4.7\% at the sentence level, and by 41.9\%, 17.5\%, and 3.9\% at the sensitive-token level. On the InternVL3.5 4B model, the corresponding improvements are 22.2\%, 18.8\%, and 3.8\% at the sentence level, and 38.4\%, 29.9\%, and 3.7\% at the sensitive-token level.

For the VQA task, our method also achieves state-of-the-art performance across all models and metrics, though the margins are generally smaller than for captioning. On the LLaVA-1.5 7B model, it improves over the best baselines by 2.6\%, 3.0\%, and 1.3\% at the sentence level, and by 13.0\%, 13.0\%, and 3.2\% at the sensitive-token level. On the Qwen2.5-VL 7B model, the corresponding improvements are 4.3\%, 3.0\%, and 1.0\% at the sentence level, and 18.6\%, 1.8\%, and 1.7\% at the sensitive-token level. On the InternVL3.5 4B model, our method achieves 9.0\%, 3.8\%, and 1.8\% improvements at the sentence level, and 30.3\%, 9.6\%, and 2.5\% at the sensitive-token level. The smaller margins in VQA reflect the fact that much of the generated output relies on reasoning and language priors rather than purely on perceptual evidence.

\begin{table*}[!t]
    \vspace{-30pt}
    \caption{Evaluation of word-level faithfulness metrics (Deletion, Insertion AUC, and Average Highest Score) and location metrics (Point Game) on the MS COCO.}\vspace{-15pt}
    \label{EAGLE:faithfulness_on_word_level}
    \begin{center}
        \resizebox{\textwidth}{!}{
            \begin{tabular}{lcl|ccc|cc|cc}
            \toprule[1.5pt]
            \multirow{2}{*}{Datasets} & \multirow{2}{*}{MLLMs} & \multirow{2}{*}{Methods} & \multicolumn{3}{c|}{Word-level Faithfulness Metrics} & \multicolumn{2}{c|}{Localization Metrics} & \multirow{2}{*}{GPU Memory ($\downarrow$)} & \multirow{2}{*}{Time Consum. ($\downarrow$)} \\ 
             &  &   & Insertion ($\uparrow$) & Deletion ($\downarrow$) & Ave. high. score ($\uparrow$) & Point Game$_{\text{bbox}}$ ($\uparrow$) & Point Game$_{\text{mask}}$ ($\uparrow$) \\ \midrule
            \multirow{16}{*}{\begin{tabular}[c]{@{}l@{}}MS COCO~\citep{lin2014microsoft}\\ (Image caption task)\end{tabular}} & \multirow{4}{*}{\begin{tabular}[c]{@{}c@{}}LLaVA-1.5 7B \citep{liu2024improved}\end{tabular}} & LLaVA-CAM~\citep{zhang2025redundancy} & 0.4063 & 0.4035 & 0.6053 & 0.2468 & 0.1168 & 36.73 GB & 17.1 s \\
            & &  iGOS++ (w/ GNC)~\citep{xing2025large} & 0.4093 & 0.3812 & 0.6084 & 0.6623 & 0.5584 & 93.12 GB & 33.5 s \\
            & & TAM~\citep{li2025token} & 0.3860 & 0.4162 & 0.5988 & 0.1818 & 0.1428 & 16.60 GB & 1.7 s \\
            & & \cellcolor[HTML]{EFEFEF}\textsc{Eagle} & \cellcolor[HTML]{EFEFEF}\textbf{0.6395} & \cellcolor[HTML]{EFEFEF}\textbf{0.2047} & \cellcolor[HTML]{EFEFEF}\textbf{0.7213} & \cellcolor[HTML]{EFEFEF}\textbf{0.8052} & \cellcolor[HTML]{EFEFEF}\textbf{0.7792} & \cellcolor[HTML]{EFEFEF}\textbf{16.31 GB} & \cellcolor[HTML]{EFEFEF}283.2 s \\
            \cmidrule(l){2-10} 
            & \multirow{4}{*}{\begin{tabular}[c]{@{}c@{}}Qwen2.5-VL 3B \citep{bai2025qwen2}\end{tabular}} & LLaVA-CAM~\citep{zhang2025redundancy} & 0.3417 & 0.4575 &  0.7263 & 0.1045 & 0.0621 & 26.01 GB & 12.8 s \\
            & &  iGOS++ (w/ GNC)~\citep{xing2025large} & 0.4141 & 0.2901 & 0.7250 & 0.5822 & 0.4967 & 58.1 GB & 49.9 s\\
            & & TAM~\citep{li2025token} &  0.5130 & 0.2797 & 0.7985 & 0.5294 & 0.4379 & 9.56 GB & 4.2 s \\
            & & \cellcolor[HTML]{EFEFEF}\textsc{Eagle} & \cellcolor[HTML]{EFEFEF}\textbf{0.7353} & \cellcolor[HTML]{EFEFEF}\textbf{0.1628} & \cellcolor[HTML]{EFEFEF}\textbf{0.8641} & \cellcolor[HTML]{EFEFEF}\textbf{0.8104} & \cellcolor[HTML]{EFEFEF}\textbf{0.7745} & \cellcolor[HTML]{EFEFEF}\textbf{9.22 GB} & \cellcolor[HTML]{EFEFEF}371.4 s \\
            \cmidrule(l){2-10} 
            & \multirow{4}{*}{\begin{tabular}[c]{@{}c@{}}Qwen2.5-VL 7B \citep{bai2025qwen2}\end{tabular}} & LLaVA-CAM~\citep{zhang2025redundancy} & 0.4170 & 0.4771 & 0.8041 & 0.2176 & 0.1428 & 44.26 GB & 24.5 s \\
            & & iGOS++ (w/ GNC)~\citep{xing2025large} & 0.4816 & 0.3478 & 0.8080 & 0.6734 & 0.5959 & 82.14 GB & 53.1 s \\
            & & TAM~\citep{li2025token} & 0.5768 & 0.3167 & 0.8240 & 0.5369 & 0.4060 & 18.75 GB & 4.9 s\\
            & & \cellcolor[HTML]{EFEFEF}\textsc{Eagle} & \cellcolor[HTML]{EFEFEF}\textbf{0.8109} & \cellcolor[HTML]{EFEFEF}\textbf{0.2127} & \cellcolor[HTML]{EFEFEF}\textbf{0.9194} & \cellcolor[HTML]{EFEFEF}\textbf{0.7785} & \cellcolor[HTML]{EFEFEF}\textbf{0.7383} & \cellcolor[HTML]{EFEFEF}\textbf{18.03 GB} & \cellcolor[HTML]{EFEFEF}435.6 s \\
            \cmidrule(l){2-10} 
            & \multirow{4}{*}{\begin{tabular}[c]{@{}c@{}}InternVL3.5 4B \citep{wang2025internvl3}\end{tabular}} & LLaVA-CAM~\citep{zhang2025redundancy} & 0.4988 & 0.5040 & 0.8588 & 0.3201 & 0.2212 & 81.84 GB & 41.3 s \\
            & & iGOS++ (w/ GNC)~\citep{xing2025large} & 0.5192 & 0.3983 & 0.8604 & 0.5775 & 0.5181 & 60.06 GB & 32.7 s \\
            & & TAM~\citep{li2025token} & 0.6317 & 0.3517 & 0.8712 & 0.5775 & 0.4653 & 14.23 GB & 4.6 s   \\
            & & \cellcolor[HTML]{EFEFEF}\textsc{Eagle} & \cellcolor[HTML]{EFEFEF}\textbf{0.8623} & \cellcolor[HTML]{EFEFEF}\textbf{0.1706} & \cellcolor[HTML]{EFEFEF}\textbf{0.9585} & \cellcolor[HTML]{EFEFEF}\textbf{0.8052} & \cellcolor[HTML]{EFEFEF}\textbf{0.7755} & \cellcolor[HTML]{EFEFEF}\textbf{7.61 GB} & \cellcolor[HTML]{EFEFEF}377.3 s \\
            \bottomrule[1.5pt]
            \end{tabular}
        }
    \end{center}
    \vspace{-20pt}
\end{table*}

In addition to higher attribution fidelity, \textsc{Eagle} demonstrates strong efficiency, requiring only 17.68 GB on Qwen2.5-VL 7B compared to 96.90 GB for IGOS++, making it practical for modern MLLMs. Overall, it provides more faithful and resource-efficient explanations than gradient-based baselines. 
Although our method is roughly ten times more time-consuming than LLaVA-CAM and I-GOS++, it is the only one that shows promising potential to meet real-world interpretability requirements.
As shown in Fig.~\ref{EGALE:sentence-level-visualization}, LLaVA-CAM often misses key regions and IGOS++ yields redundant maps, while our method highlights critical regions that align closely with visually grounded tokens, producing concise and human-consistent explanations.

\subsection{Experiments on Word-level Explanations}

Next, we evaluate the ability of the proposed attribution method to provide word-level explanations. Specifically, we use samples with object bounding box annotations from the MS COCO dataset to verify whether the objects mentioned in image captions are accurately grounded in the visual input. We also include TAM~\citep{li2025token} as an additional baseline, since it is particularly effective at explaining object localization.

Table~\ref{EAGLE:faithfulness_on_word_level} reports the results of faithfulness and localization evaluations, where our method consistently achieves state-of-the-art performance across all models and metrics. For faithfulness, on the LLaVA-1.5 7B model, it surpasses the strongest baseline by 56.2\%, 46.3\%, and 19.3\% in insertion, deletion, and average highest score, respectively. On the Qwen2.5-VL 7B model, the corresponding improvements are 40.6\%, 10.4\%, and 11.6\%, while on the InternVL3.5 4B model, they are 36.5\%, 51.5\%, and 10.0\%. We also observe that TAM performs well only on stronger MLLMs such as Qwen2.5-VL and InternVL3.5, since it relies solely on activation maps rather than capturing strong causal relationships. In contrast, our method is broadly applicable across models and can faithfully explain word-level decisions even for LLaVA-1.5.

\begin{figure}
    \centering
    \includegraphics[width=0.48\textwidth]{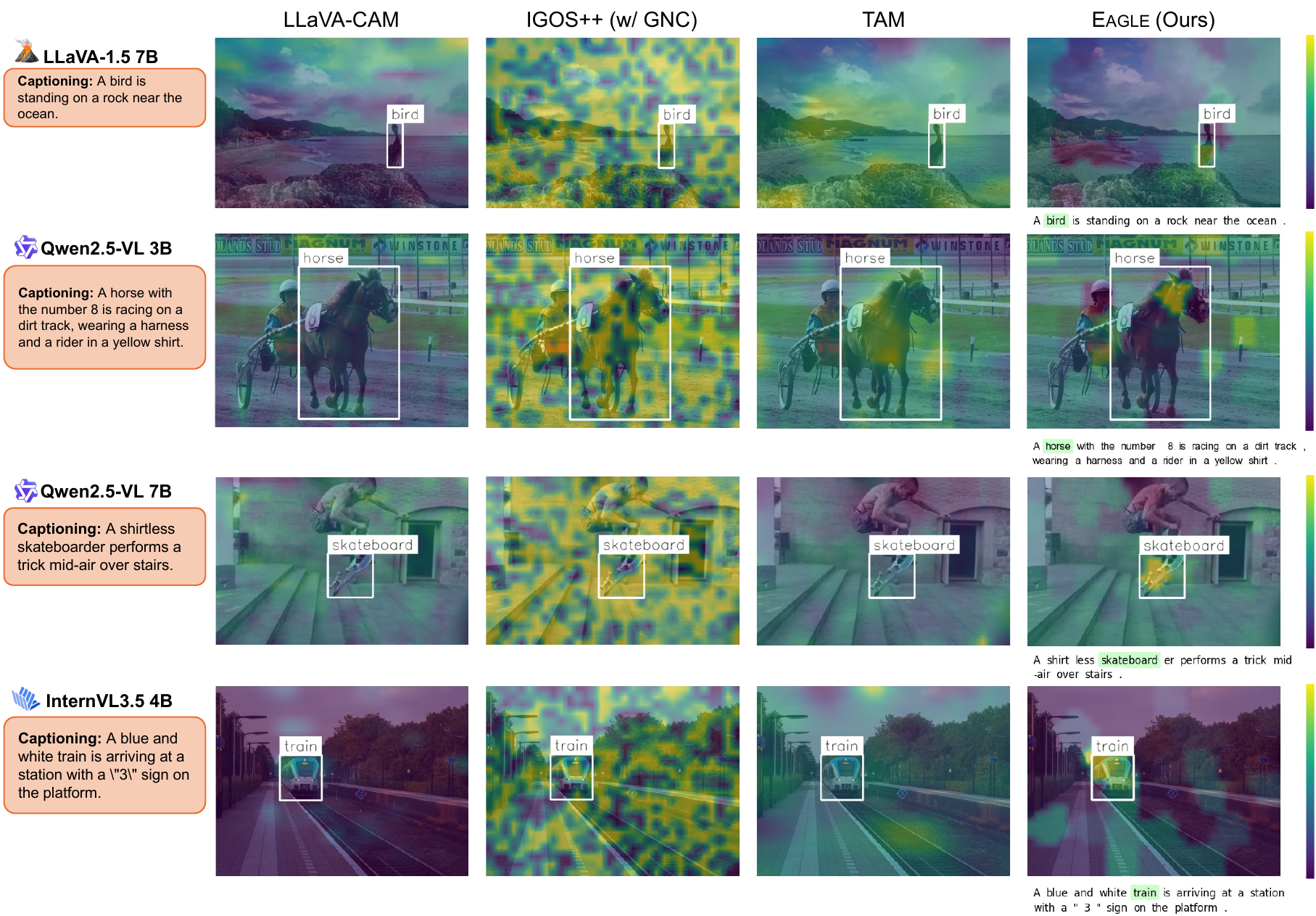}\vspace{-8pt}
    \caption{Visualization of word-level explanation results for LLaVA-1.5, Qwen2.5-VL, and InternVL3.5 on the MS COCO datasets.} 
    \label{EGALE:word-level-visualization}
    \vspace{-15pt}
\end{figure}

For localization, our method achieves the best Pointing Game results under both box- and mask-level settings, confirming that predictions are grounded in specific objects. While TAM performs well on stronger models but poorly on LLaVA-1.5, IGOS++ gains from overly redundant maps. In contrast, our method yields sparse yet focused highlights that more accurately localize the objects mentioned in captions (Fig.~\ref{EGALE:word-level-visualization}). Although TAM~\cite{li2025token} is much faster in attribution time, it heavily relies on the MLLM’s activation maps. As a result, while it performs well on Qwen2.5-VL and InternVL 3.5, its effectiveness drops sharply on LLaVA-1.5. In contrast, our method does not depend on model-specific internal activations and therefore remains consistently superior.

\begin{table*}[!t]
    \vspace{-30pt}
    \caption{Evaluation of faithfulness metrics and correction-oriented metrics on hallucination interpretation.}\vspace{-12pt}
    \label{EAGLE:faithfulness_on_hallucination}
    \begin{center}
        \resizebox{0.9 \textwidth}{!}{
            \begin{tabular}{lcl|ccc|cc|c}
            \toprule[1.5pt]
            \multirow{2}{*}{Datasets} & \multirow{2}{*}{MLLMs} & \multirow{2}{*}{Methods} & \multicolumn{3}{c|}{Faithfulness Metrics} & \multicolumn{2}{c|}{Correction-oriented Metrics} & \multirow{2}{*}{GPU Memory ($\downarrow$)} \\ 
             &  &   & Insertion ($\uparrow$) & Deletion ($\downarrow$) & Ave. high. score ($\uparrow$) & AMCR ($\downarrow$) & CSR@10\% ($\uparrow$) & \\ \midrule
            \multirow{16}{*}{\begin{tabular}[c]{@{}l@{}}RePOPE~\citep{neuhaus2025repope}\\ (Object Hallucination Benchmark)\end{tabular}} & \multirow{4}{*}{\begin{tabular}[c]{@{}c@{}}LLaVA-1.5 7B \citep{liu2024improved}\end{tabular}} & LLaVA-CAM~\citep{zhang2025redundancy} & 0.4095 & 0.4191 & 0.6596 & 0.5613 & 19.70\% & 37.07 GB  \\
            & & iGOS++ (w/ GNC)~\citep{xing2025large} & 0.4232 & 0.4182 & 0.6794 & 0.4770 & 37.50\% & 93.88 GB \\
            & & TAM~\citep{li2025token} & 0.4168 & 0.4166 & 0.6705 & 0.5826 & 18.75\% & 16.59 GB \\
            & & \cellcolor[HTML]{EFEFEF}\textsc{Eagle} & \cellcolor[HTML]{EFEFEF}\textbf{0.6999} & \cellcolor[HTML]{EFEFEF}\textbf{0.2652} & \cellcolor[HTML]{EFEFEF}\textbf{0.7877} & \cellcolor[HTML]{EFEFEF}\textbf{0.0844} & \cellcolor[HTML]{EFEFEF}\textbf{77.50\%} & \cellcolor[HTML]{EFEFEF}\textbf{16.04 GB} \\
            \cmidrule(l){2-9} 
            & \multirow{4}{*}{\begin{tabular}[c]{@{}c@{}}Qwen2.5-VL 3B \citep{bai2025qwen2}\end{tabular}} & LLaVA-CAM~\citep{zhang2025redundancy} &  0.3994  & 0.3783 & 0.6992 & 0.4555 & 42.21\% & 27.10 GB  \\
            & & iGOS++ (w/ GNC)~\citep{xing2025large} & 0.4056 & 0.4471 & 0.7235 & 0.4461 & 37.57\% & 35.16 GB \\
            & & TAM~\citep{li2025token} & 0.3905 & 0.4090 & 0.6900 & 0.4747 & 29.75\% & 9.66 GB \\
            & & \cellcolor[HTML]{EFEFEF}\textsc{Eagle} & \cellcolor[HTML]{EFEFEF}\textbf{0.7568} & \cellcolor[HTML]{EFEFEF}\textbf{0.1610} & \cellcolor[HTML]{EFEFEF}\textbf{0.8717} & \cellcolor[HTML]{EFEFEF}\textbf{0.0849} & \cellcolor[HTML]{EFEFEF}\textbf{80.41\%} & \cellcolor[HTML]{EFEFEF}\textbf{9.20 GB} \\
            \cmidrule(l){2-9} 
            & \multirow{4}{*}{\begin{tabular}[c]{@{}c@{}}Qwen2.5-VL 7B \citep{bai2025qwen2}\end{tabular}} & LLaVA-CAM~\citep{zhang2025redundancy} & 0.2444 & 0.2901 & 0.5898 & 0.6620 & 35.37\% & 45.05 GB \\
            & & iGOS++ (w/ GNC)~\citep{xing2025large} & 0.3017 & 0.3330 & 0.7125 & 0.5357 & 32.41\% & 70.86 GB \\
            & & TAM~\citep{li2025token} & 0.2717 & 0.3177 & 0.6792 & 0.5844 & 22.45\% & 18.57 GB \\
            & & \cellcolor[HTML]{EFEFEF}\textsc{Eagle} & \cellcolor[HTML]{EFEFEF}\textbf{0.7987} & \cellcolor[HTML]{EFEFEF} \textbf{0.0331} & \cellcolor[HTML]{EFEFEF} \textbf{0.9381} & \cellcolor[HTML]{EFEFEF} \textbf{0.1442} & \cellcolor[HTML]{EFEFEF} \textbf{73.94\%} & \cellcolor[HTML]{EFEFEF}\textbf{18.26 GB} \\
            \cmidrule(l){2-9} 
            & \multirow{4}{*}{\begin{tabular}[c]{@{}c@{}}InternVL3.5 4B \citep{wang2025internvl3}\end{tabular}} & LLaVA-CAM~\citep{zhang2025redundancy} & 0.4079 & 0.3733 & 0.9296 & 0.4078 & 36.43\% & 87.93 GB \\
            & & iGOS++ (w/ GNC)~\citep{xing2025large} & 0.3651 & 0.4556 & 0.9393 & 0.4299 & 38.76\% & 66.26 GB \\
            & & TAM~\citep{li2025token} & 0.3794 & 0.4221 & 0.9115 & 0.4801 & 28.57\% & 14.04 GB \\
            & & \cellcolor[HTML]{EFEFEF}\textsc{Eagle} & \cellcolor[HTML]{EFEFEF}\textbf{0.9114} & \cellcolor[HTML]{EFEFEF}\textbf{0.0440} & \cellcolor[HTML]{EFEFEF}\textbf{0.9941} & \cellcolor[HTML]{EFEFEF}\textbf{0.0676} & \cellcolor[HTML]{EFEFEF}\textbf{80.00\%} & \cellcolor[HTML]{EFEFEF} \textbf{12.31 GB} \\
            \bottomrule[1.5pt]
            \end{tabular}
        }
    \end{center}
    \vspace{-20pt}
\end{table*}

\subsection{Interpreting Object Hallucinations}

We next apply our interpretable algorithm to analyze the causes of hallucinations in MLLMs. Experiments are conducted on the object hallucination benchmark RePOPE~\citep{neuhaus2025repope}. Our focus is on samples where the MLLM makes prediction errors, particularly in cases where the model incorrectly answers "no" instead of "yes," and vice versa. The primary aim is not to mitigate hallucinations, but to explain why they occur, specifically, by identifying the image regions that trigger hallucinations. Assuming that hallucinations have already been identified, we aim to pinpoint which image regions are responsible for these errors and to evaluate whether blocking these regions can reduce the hallucinations. Practically, we attribute the first token of the answer, focusing on the vocabulary IDs 'Yes' and 'No'. For example, if the model incorrectly outputs 'Yes', the attribution is computed with respect to 'No', providing a counterfactual perspective on which regions would support the correct response.

Table~\ref{EAGLE:faithfulness_on_hallucination} reports the results of attributing hallucinations to specific input regions. On the LLaVA-1.5 7B model, our method improves over the strongest baseline by 65.4\%, 36.3\%, and 15.9\% in insertion, deletion, and average highest score, respectively. On the Qwen2.5-VL 7B model, the gains are even larger, reaching 164.7\%, 88.6\%, and 31.7\%, while on the InternVL3.5 4B model, the improvements are 123.4\%, 88.2\%, and 5.8\%. These substantial margins highlight the strength of our approach in faithfully uncovering the input regions responsible for hallucinated predictions and in explaining the underlying causes of incorrect decisions, revealing not only where the model looked, but also why it went wrong.

\begin{figure*}
    \includegraphics[width=\textwidth]{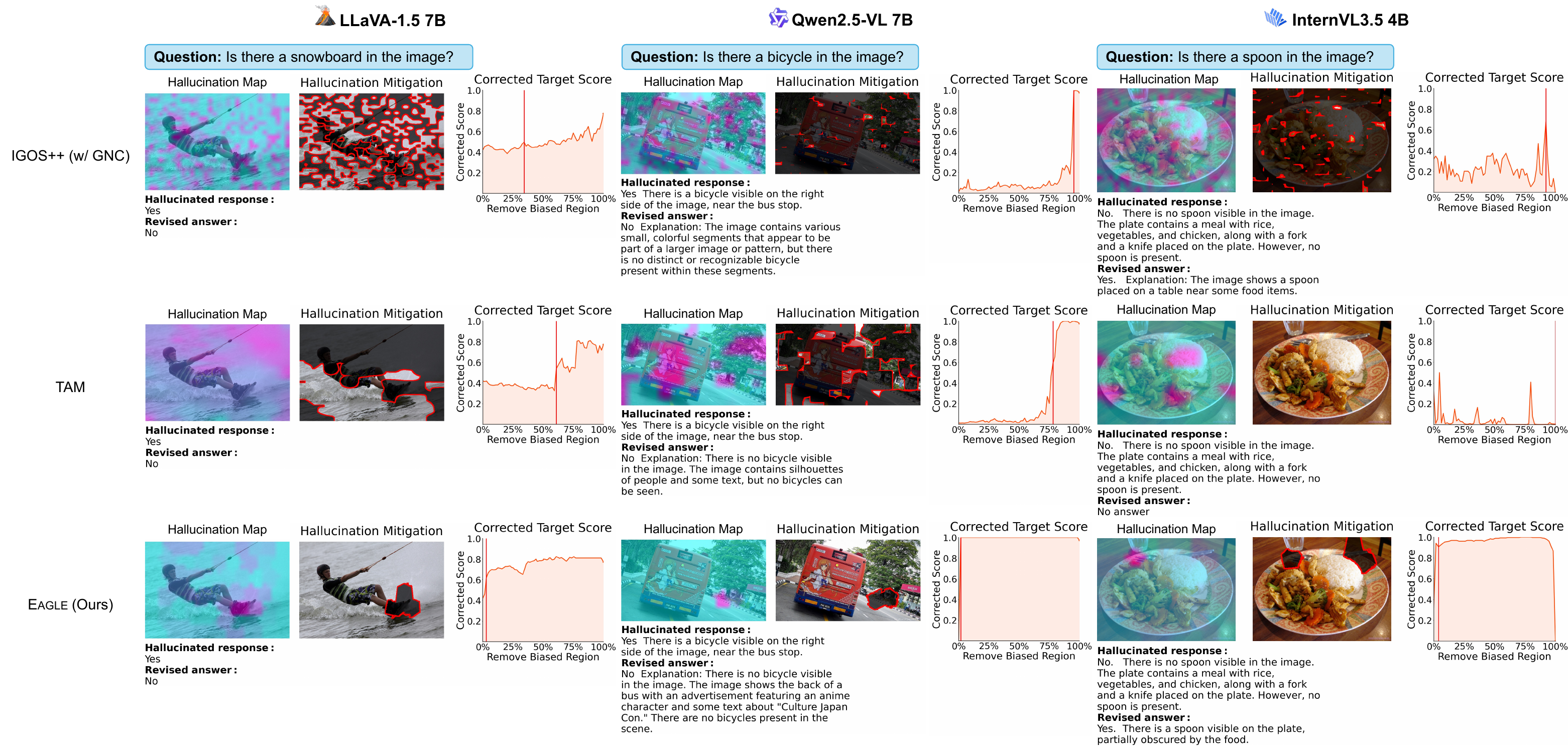}\vspace{-8pt}
    \caption{Hallucination attribution on RePOPE. Our method produces sparse, focused maps that more accurately reveal regions responsible for hallucinated outputs, compared with IGOS++ and TAM.} 
    \label{EGALE:hallucination-visualization}
    \vspace{-15pt}
\end{figure*}

Next, we examine whether hallucinations can be eliminated by progressively removing the responsible regions. Instead of relying on logits, we evaluate direct model outputs (Yes or No with the corresponding rationale) using correction-oriented metrics. On the LLaVA-1.5 7B model, our method surpasses the strongest baseline by 82.3\% and 106.6\% in Average Minimal Correction Region (AMCR) and Correction Success Rate under Budget (CSR@10\%), respectively. On the Qwen2.5-VL 7B model, the improvements are 73.1\% and 109.0\%, and on the InternVL3.5 4B model they are 83.4\% and 106.4\%. These results show that removing only a small portion of the input is sufficient to eliminate hallucinations, demonstrating the effectiveness of our attribution approach.

\begin{table}[!t]
    \caption{Ablation of objective function components on Qwen2.5-VL 7B for MS COCO captioning.}\vspace{-12pt}
    \label{EAGLE:ablation_function}
    \begin{center}
        \resizebox{0.35\textwidth}{!}{
            \begin{tabular}{cc|ccc}
                \toprule
                Insight & Necessity & \multicolumn{3}{c}{Faithfulness Metrics}  \\ 
                (Eq.~\ref{EAGLE:insight}) & (Eq.~\ref{EAGLE:necessity}) & Ins. ($\uparrow$)  & Del. ($\downarrow$) & Avg. High ($\uparrow$) \\ 
                \midrule
                \ding{55} & \ding{51}  & 0.6176 & \underline{0.4613} &  0.7282 \\
                \ding{51} & \ding{55}  & \underline{0.6981} & 0.5253 & \underline{0.7566} \\
                \ding{51} & \ding{51}   & \textbf{0.7006} & \textbf{0.4597} & \textbf{0.7578} \\  
                \bottomrule
            \end{tabular}
        }
    \end{center}
    \vspace{-20pt}
\end{table}

Fig.~\ref{EGALE:hallucination-visualization} visualizes the results, including the Hallucination Map, where highlighted purple regions indicate areas prone to hallucinations identified by our method. Hallucination Mitigation denotes the minimal region that must be removed to eliminate hallucinations. The curve illustrates changes in the logit of the ground-truth token as hallucination-prone regions are progressively deleted, with the red line marking the deletion point determined by Hallucination Mitigation. Our method rapidly localizes regions that cause hallucinations, while TAM and IGOS++ produce diffuse maps. On LLaVA-1.5, it attributes the false detection of a snowboard to a surfboard, highlighting confusion between similar objects. InternVL3.5 fails to recognize a spoon that is partially occluded by a fork. By precisely attributing and removing the fork head, our method enables the model to correctly identify the spoon, revealing its limited ability to disambiguate overlapping objects.

\subsection{Ablation Study} 



We conduct ablations on the MS COCO captioning task with Qwen2.5-VL 7B to evaluate both the objective function design and the impact of subregion partitioning. As shown in Table~\ref{EAGLE:ablation_function}, only the joint use of the Insight and Necessity Scores consistently improves all faithfulness metrics, demonstrating their complementary effects. Table~\ref{EAGLE:ablation_number} further shows that finer image partitions generally enhance faithfulness, though at the expense of increased attribution time, suggesting the importance of developing more scalable attribution strategies in future work.

\begin{table}[!t]
    \caption{Ablation of subregion number on Qwen2.5-VL 7B for MS COCO captioning.}\vspace{-12pt}
    \label{EAGLE:ablation_number}
    \begin{center}
        \resizebox{0.3\textwidth}{!}{
            \begin{tabular}{c|ccc}
            \toprule
            \multirow{2}{*}{Number} & \multicolumn{3}{c}{Faithfulness Metrics} \\ 
                                     & Ins. ($\uparrow$) & Del. ($\downarrow$) & Avg. High ($\uparrow$) \\ 
            \midrule
            36 & 0.6869 & 0.4587 & 0.7452 \\
            50 & 0.6901 & \textbf{0.4514} & 0.7482 \\
            64 & \textbf{0.7006} & 0.4597 & \textbf{0.7578} \\
            \bottomrule
            \end{tabular}
        }
    \end{center}
    \vspace{-20pt}
\end{table}

\section{Conclusion}

In this paper, we present \textsc{Eagle}, a black-box attribution framework for autoregressive MLLMs. By unifying sufficiency and indispensability in a submodular-inspired objective, \textsc{Eagle} faithfully explains token generation, revealing both \emph{where} models attend and \emph{what} they rely on. Experiments across diverse models and datasets show clear gains in faithfulness, localization, and hallucination diagnosis. Moreover, \textsc{Eagle} pinpoints the minimal visual factors that give rise to hallucinations, providing an effective, interpretation-focused account of their causes.

\section*{Acknowledgements} This work was supported by the Guangdong Major Project of Basic Research (No. 2023B0303000010), National Natural Science Foundation of China (No. 62372448, 62132006, U2541229), and Shenzhen Science and Technology Program (No. CJGJZD20240729141505007).

{
    \small
    \bibliographystyle{ieeenat_fullname}
    \bibliography{main}
}

\clearpage
\setcounter{page}{1}
\maketitlesupplementary


\section{\textsc{Eagle} Algorithm}

The detailed calculation process of the proposed EAGLE algorithm is outlined below.

\begin{algorithm}[h]
    \caption{\textsc{Eagle}: Explaining Autoregressive Generation by Language priors or Evidence in multimodal large language models (MLLMs)}
    \label{EAGLE:alg}
    \KwIn{Image $\mathbf{I} \in \mathbb{R}^{h \times w \times 3}$, partitioning algorithm $\texttt{Div}(\cdot)$, prompt $\texttt{Prompt}$, generated sequence $\mathbf{y}$, target token positions $T$, vocabulary indices $\mathcal{V}$.}
    \KwOut{Ordered subset $\pi$, saliency map $\mathcal{A} \in \mathbb{R}^{h \times w}$, influence scores $I_t$.}
    $V \gets \texttt{Div}(\mathbf{I})$\;
    $\pi \gets \varnothing$ \Comment*[r]{Initialize ordered subset}
    $\mathcal{A}_1 \gets 0$\;
    \For{$i=1$ \KwTo $|V|$}{
        $S_d \gets V \setminus S$\;
        $\alpha \gets \arg\max_{\alpha \in S_d}\,\mathcal{F}\!\left(\pi \cup \{\alpha\}\right)$\;
        $\pi \gets \pi \,\|\, \{\alpha\}$\;
        \uIf{$i>1$}{
            $\mathcal{A}_i \gets \mathcal{A}_{i-1} - \big| \mathcal{F}(\pi_{:i}) - \mathcal{F}(\pi_{:i-1}) \big|$ \Comment*[r]{Saliency update}
        }
    }
    \For{$i=1$ \KwTo $|T|$}{    
        $s_{\max} \gets \max_{1 \le j \le |\pi|} p\!\left(y_{t_i} = v_i \mid \pi_{:j}, \texttt{Prompt}, \mathbf{y}_{<t_i}\right)$\;
        $I_{t_i} \gets \sum_{r=1}^{|\pi|} \Big( s_{\max} - p\!\left(y_{t_i} = v_i \mid \pi_{:r}, \texttt{Prompt}, \mathbf{y}_{<t_i}\right) \Big)$ \Comment*[r]{Language prior vs. perception evidence}
    }
    
    \Return $\pi,\, \text{norm}(\mathcal{A}),\, \text{norm}(I_t)$
\end{algorithm}

\section{Additional Experimental Details}

For the image captioning task on MS COCO, the prompt used for all MLLMs is:

\texttt{Describe the image in one factual English sentence of no more than 20 words. Do not include information that is not clearly visible.}

For the hallucination detection task on RePOPE, the prompt used is:



{\small
\begin{verbatim}
You are asked a visual question answering task. 
First, answer strictly with "Yes" or "No". 
Then, provide a short explanation if necessary.

Question: {question}
Answer:
\end{verbatim}
}

\section{Limitations and Future Works.} 

Despite its effectiveness, our work has two main limitations. First, the iterative subset selection and greedy search limit scalability compared with lightweight visualization methods. Nevertheless, our approach provides a promising interpretable pathway and clarifies a potential upper bound of attribution for MLLMs; in future work, we will design more efficient attribution algorithms. Second, the framework focuses on hallucination explanation and partial mitigation, leaving proactive prevention unexplored. In future work, we will leverage explanations to automatically detect and mitigate hallucinations and, once failure modes are identified, develop data-/parameter-efficient methods for minimal-cost model repair. 



\section{Additional Qualitative Results}

In this appendix, we provide extended qualitative visualizations that complement the main findings in Fig.~\ref{EGALE:sentence-level-visualization}, Fig.~\ref{EGALE:word-level-visualization}, and Fig.~\ref{EGALE:hallucination-visualization}. These supplementary results aim to offer a finer-grained perspective on how competing attribution methods and our proposed approach behave across diverse settings. Specifically, we present: (i) sentence-level explanations on both MS COCO and MMVP, (ii) word-level explanations on MS COCO, and (iii) hallucination attribution visualizations on additional samples. Collectively, these results provide deeper insights into the consistency, precision, and interpretability of our method.

\subsection{Sentence-level Explanations on MS COCO and MMVP}

\begin{figure*}[t]
    \centering
    \includegraphics[width=0.9\textwidth]{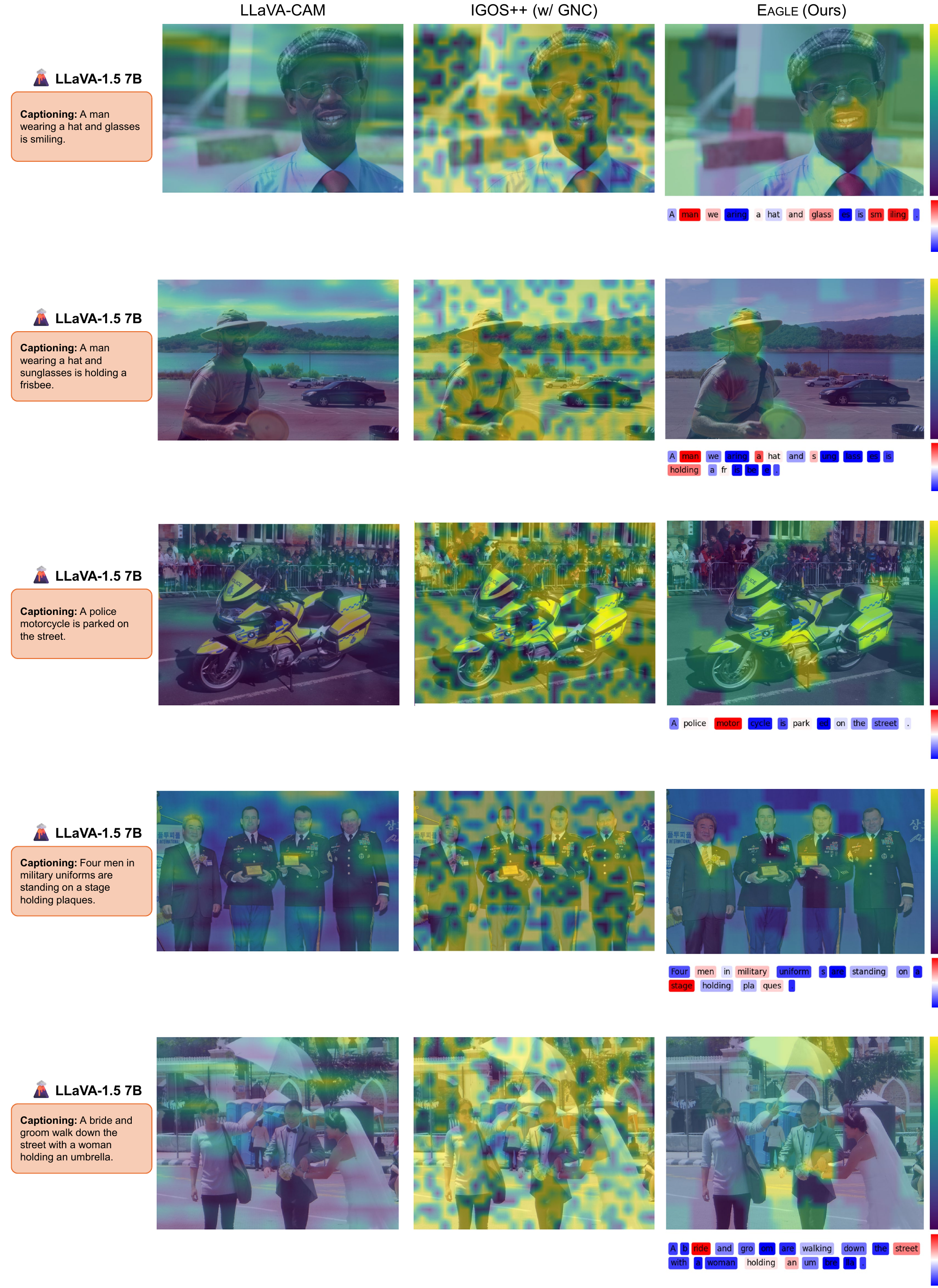}
    \caption{Sentence-level explanation results for \textbf{LLaVA-1.5} on the MS COCO dataset. Our method consistently identifies semantically critical regions that align with highlighted tokens in the caption, while baseline methods either fail to capture relevant areas (LLaVA-CAM) or over-highlight irrelevant background regions (IGOS++).}
    \label{app:llava_caption}
\end{figure*}

\begin{figure*}[t]
    \centering
    \includegraphics[width=0.9\textwidth]{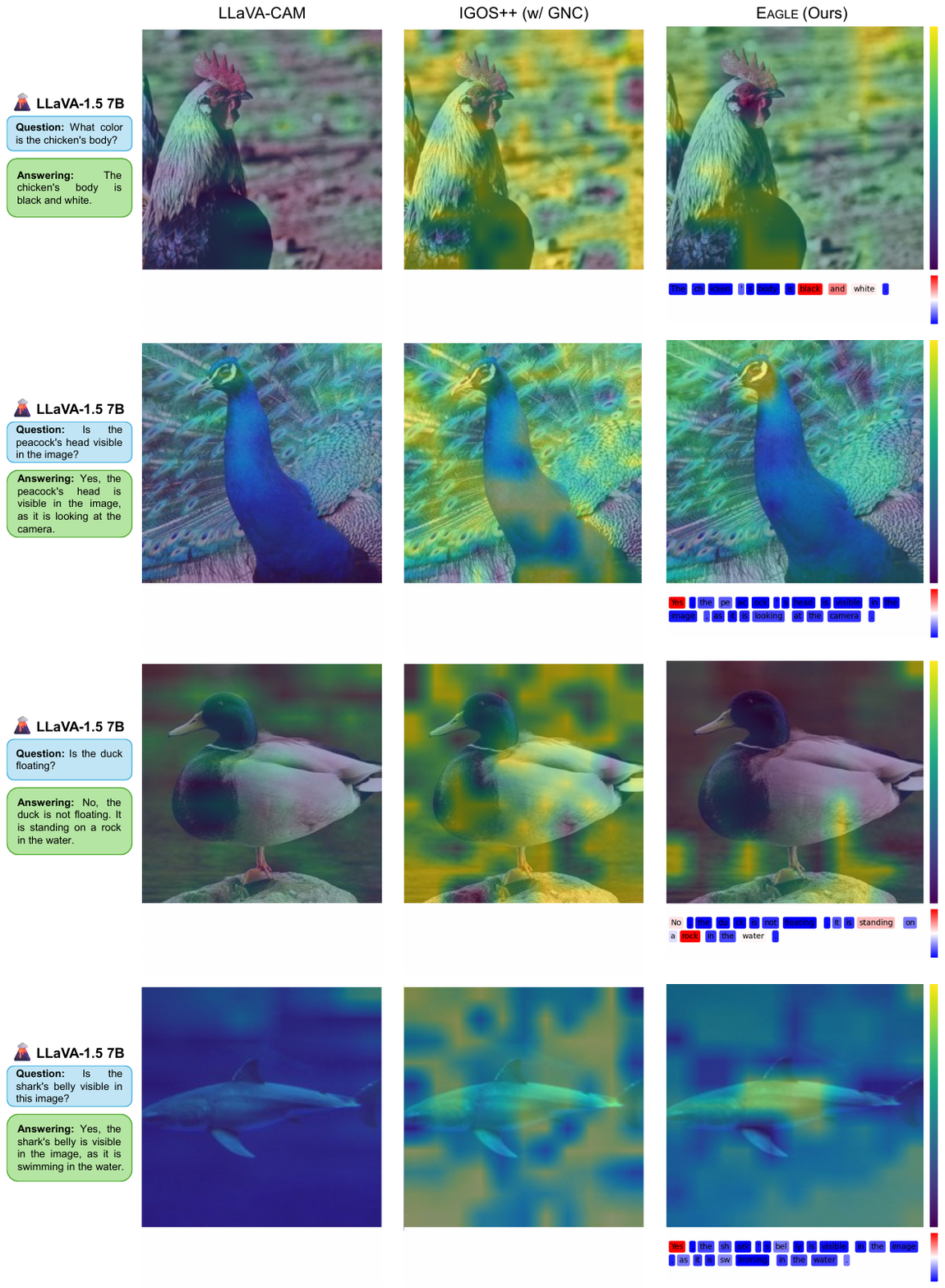}
    \caption{Sentence-level explanation results for \textbf{LLaVA-1.5} on the MMVP dataset. Compared to the baselines, our method highlights regions that are directly related to the VQA queries, resulting in explanations that are more interpretable and trustworthy.}
    \label{app:llava_vqa}
\end{figure*}

As shown in Fig.~\ref{app:llava_caption} and Fig.~\ref{app:llava_vqa}, our method produces faithful explanations for \textbf{LLaVA-1.5} by tightly aligning highlighted regions with relevant caption tokens (e.g., ``smiling,'' ``hat,'' ``motor'') or VQA queries (e.g., ``Is the shark's belly visible?''). In contrast, LLaVA-CAM often distributes attention diffusely across the scene, while IGOS++ over-activates irrelevant background regions.

\begin{figure*}[t]
    \centering
    \includegraphics[width=0.9\textwidth]{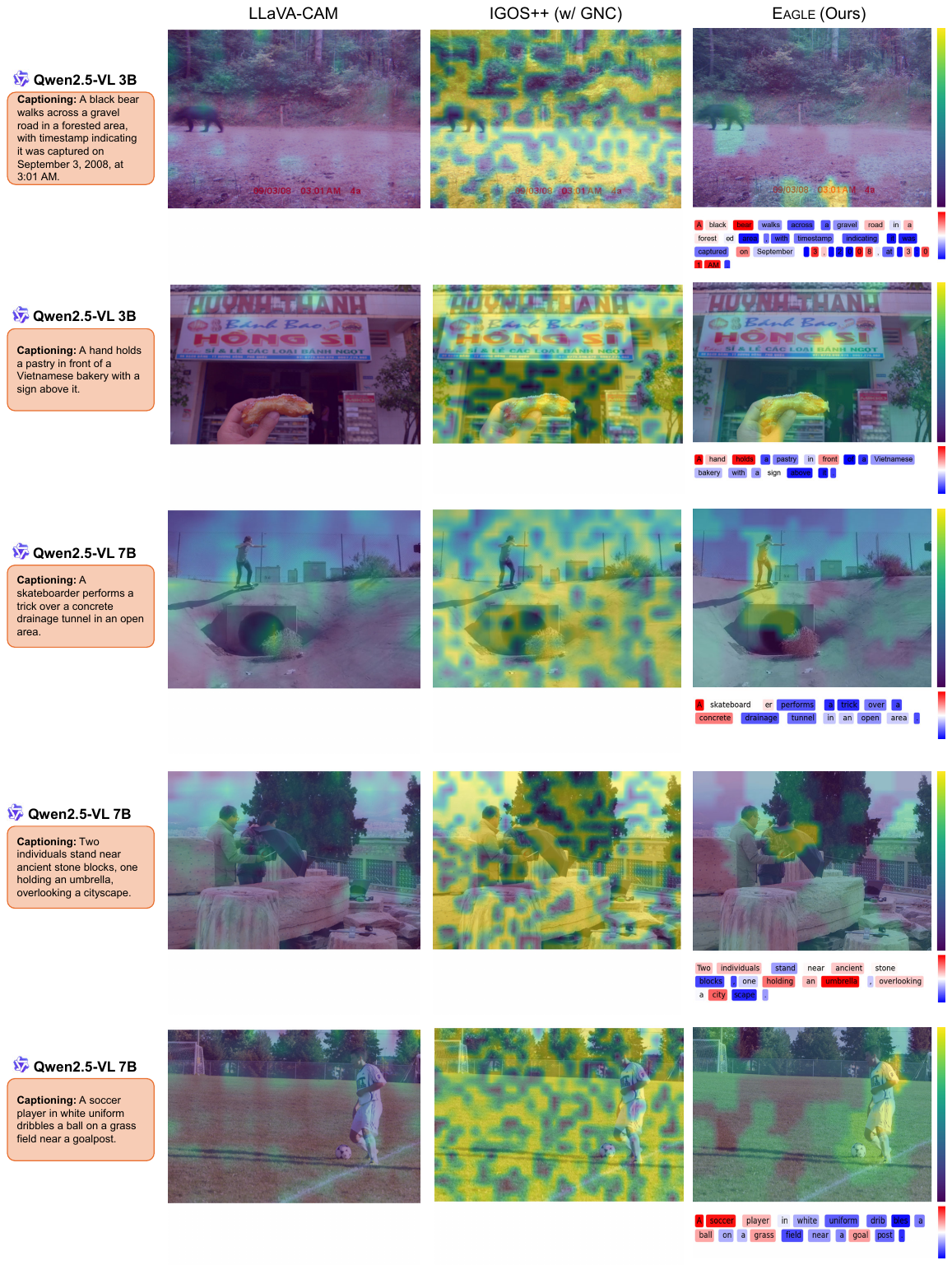}
    \caption{Sentence-level explanation results for \textbf{Qwen2.5-VL} on the MS COCO dataset. Our method highlights critical objects with strong correspondence to the generated captions, reducing redundancy in comparison to IGOS++.}
    \label{app:qwen_caption}
\end{figure*}

\begin{figure*}[t]
    \centering
    \includegraphics[width=0.9\textwidth]{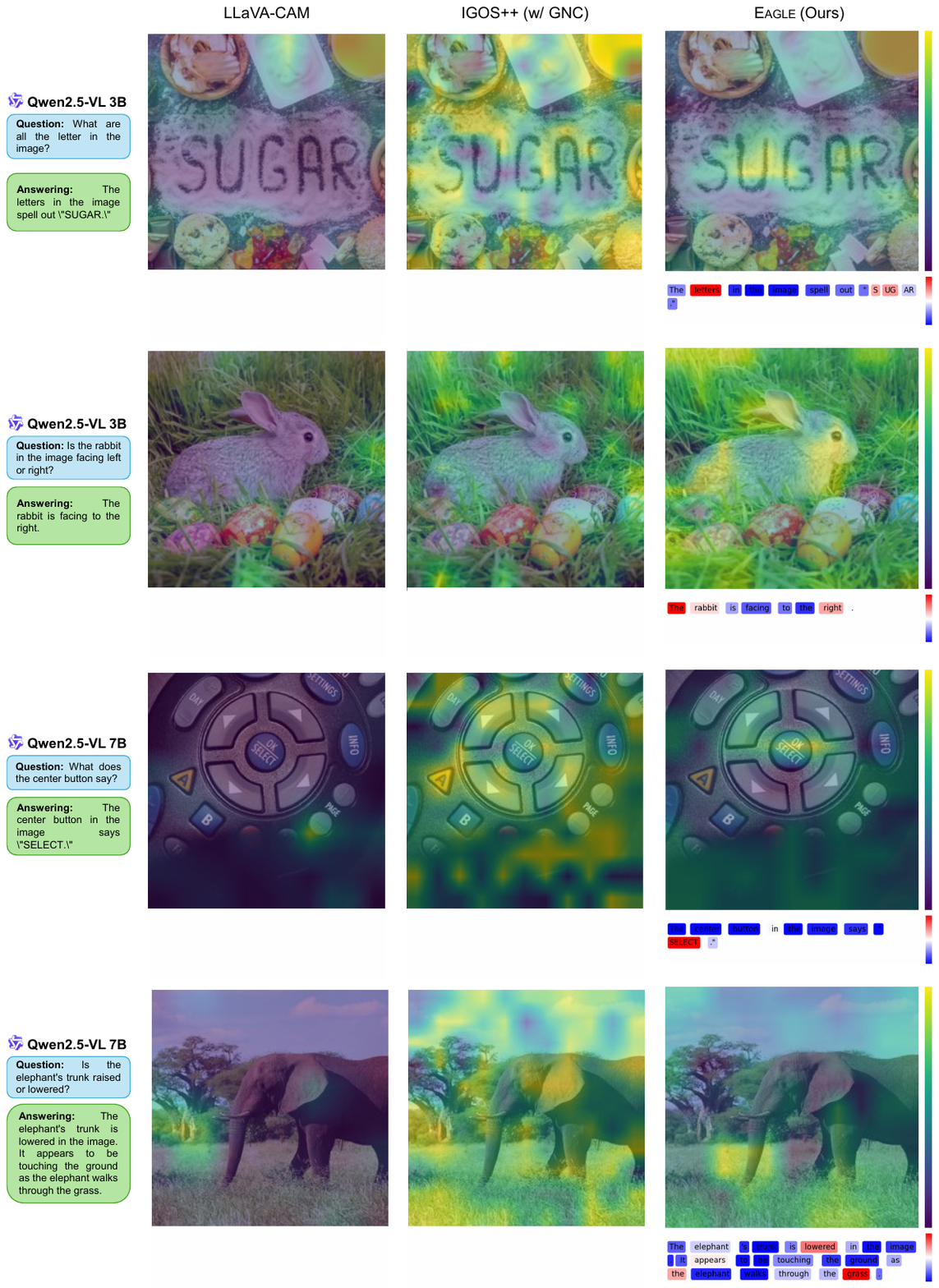}
    \caption{Sentence-level explanation results for \textbf{Qwen2.5-VL} on the MMVP dataset. Our method improves alignment between highlighted visual regions and VQA-relevant words, enhancing interpretability.}
    \label{app:qwen_vqa}
\end{figure*}

For \textbf{Qwen2.5-VL}, Fig.~\ref{app:qwen_caption} and Fig.~\ref{app:qwen_vqa} show that our method generates concise and semantically meaningful attribution maps. For example, in captions mentioning multiple objects, our approach selectively highlights the relevant ones while avoiding redundancy. In VQA tasks, it accurately isolates queried entities such as a remote button, whereas baselines either miss the target or introduce noise.

\begin{figure*}[t]
    \centering
    \includegraphics[width=0.9\textwidth]{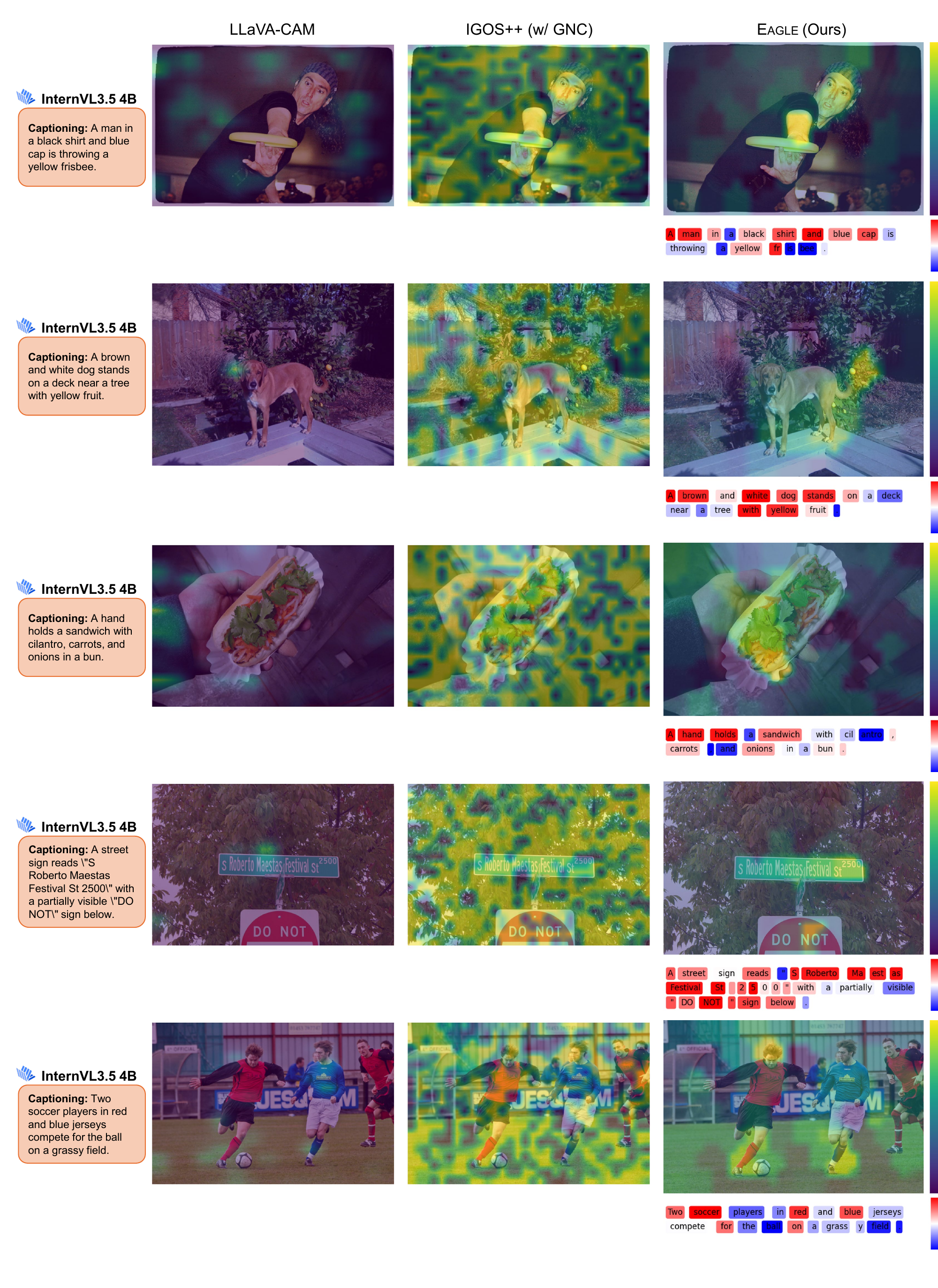}
    \caption{Sentence-level explanation results for \textbf{InternVL3.5} on the MS COCO dataset. Our method captures object-centric regions more consistently than baseline methods.}
    \label{app:internvl_caption}
\end{figure*}

\begin{figure*}[t]
    \centering
    \includegraphics[width=0.9\textwidth]{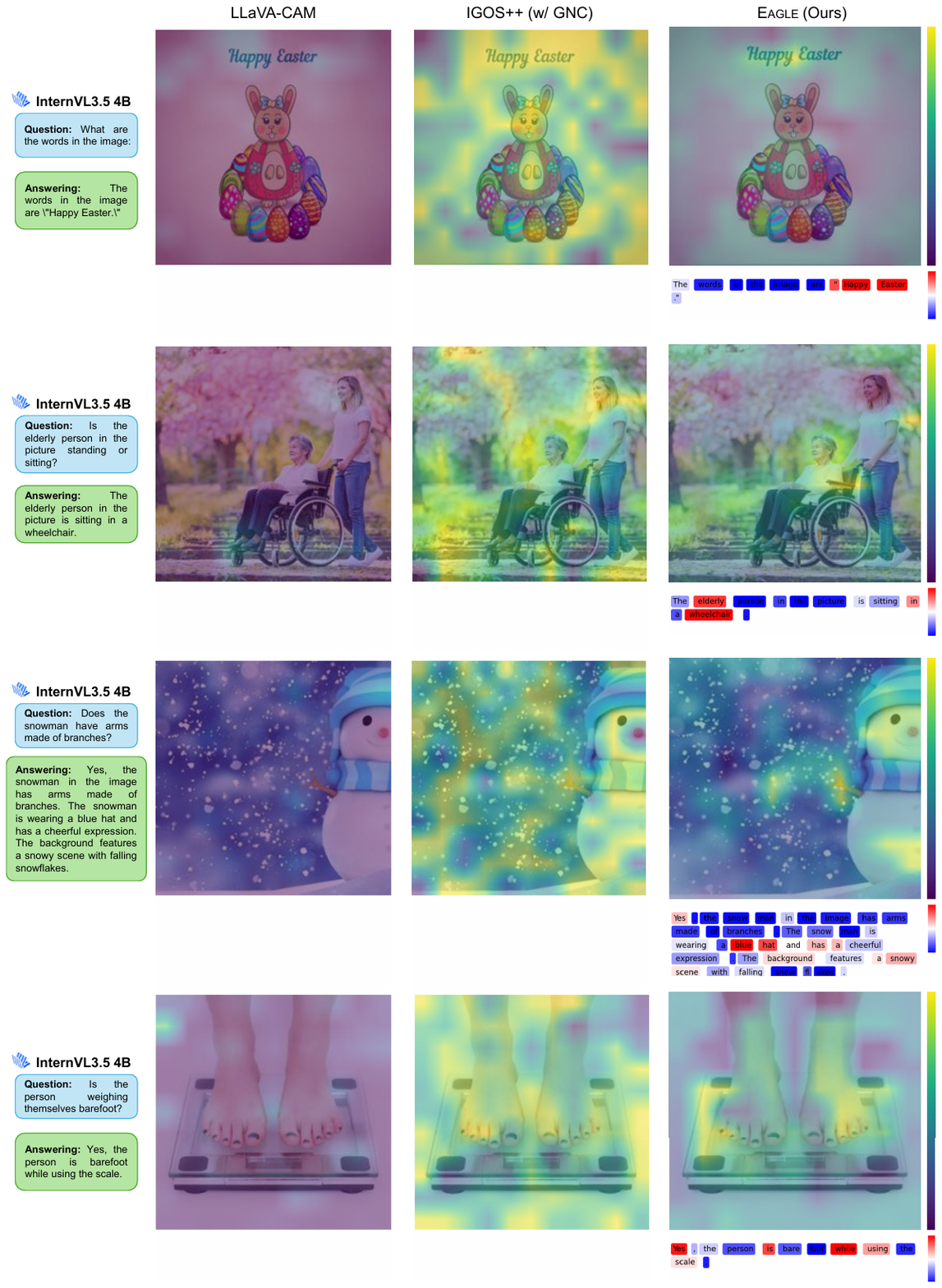}
    \caption{Sentence-level explanation results for \textbf{InternVL3.5} on the MMVP dataset. Our approach ensures strong consistency between highlighted evidence and the VQA queries.}
    \label{app:internvl_vqa}
\end{figure*}

Similarly, for \textbf{InternVL3.5} (Fig.~\ref{app:internvl_caption}, Fig.~\ref{app:internvl_vqa}), our method highlights precise object-centric regions corresponding to key caption tokens (e.g., ``sandwich,'' ``frisbee'') and VQA queries (e.g., ``Does the snowman have arms made of branches?''). Baseline methods either scatter attention broadly or fail to capture the queried object, reducing interpretability. These results collectively demonstrate that our approach consistently improves faithfulness and transparency across different models and datasets.

\subsection{Word-level Explanations on MS COCO}

\begin{figure*}[t]
    \centering
    \includegraphics[width=\textwidth]{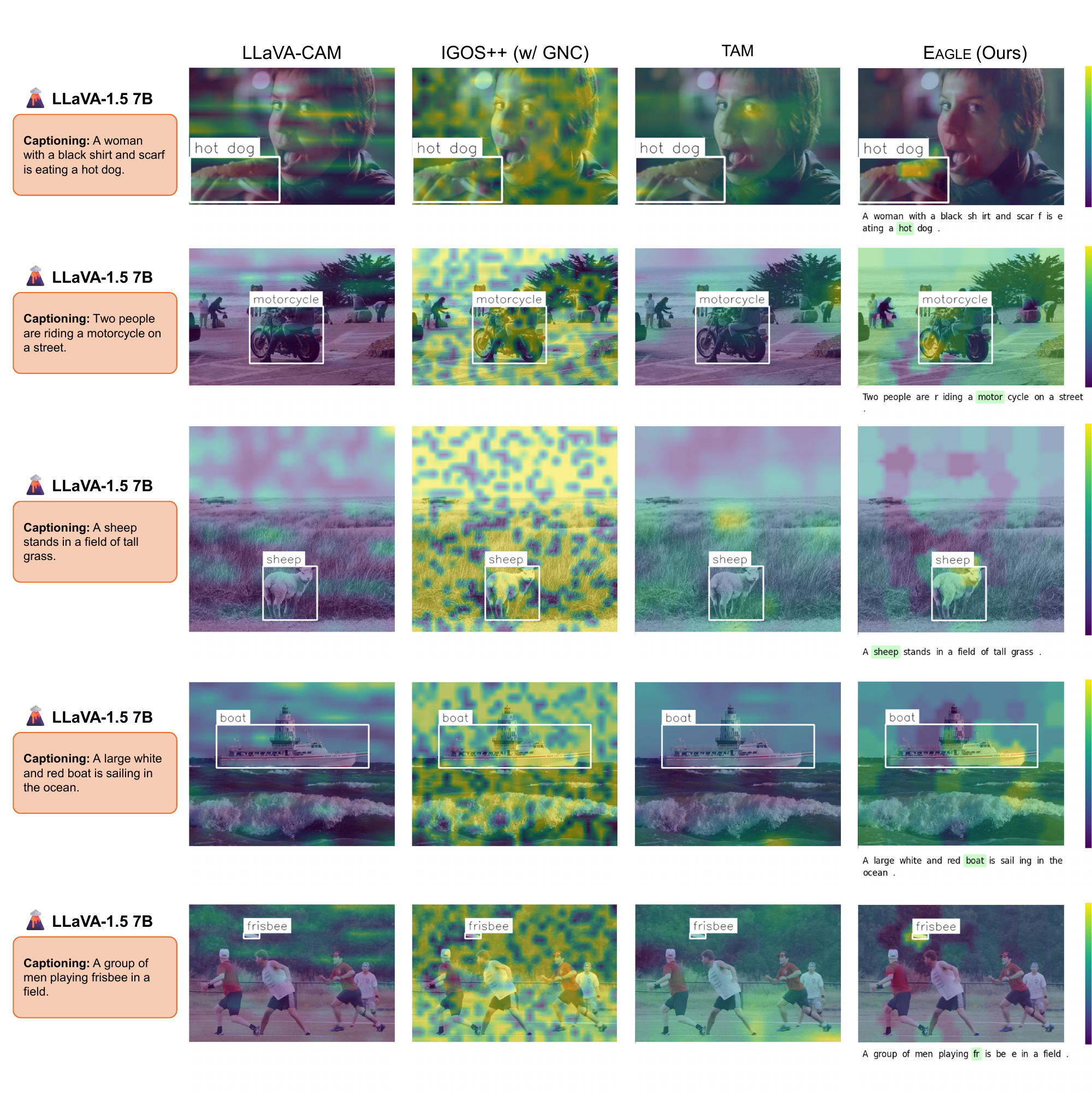}
    \caption{Word-level explanation results for \textbf{LLaVA-1.5} on the MS COCO dataset. Bounding box overlays show that our method provides sparse yet highly accurate localization.}
    \label{app:llava_object}
\end{figure*}

\begin{figure*}[t]
    \centering
    \includegraphics[width=\textwidth]{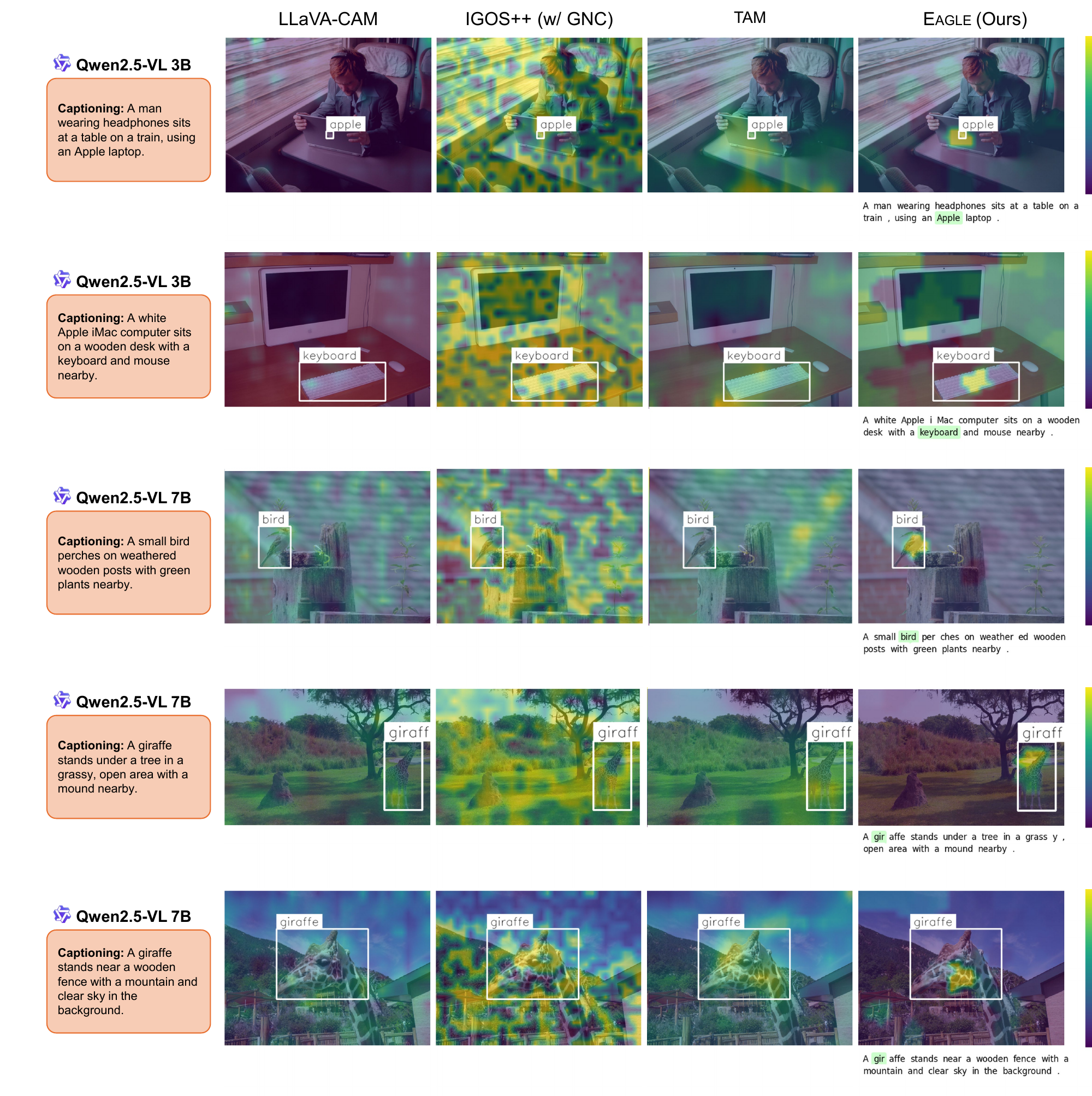}
    \caption{Word-level explanation results for \textbf{Qwen2.5-VL} on the MS COCO dataset. Our method produces localized attribution maps with high correspondence to ground-truth bounding boxes.}
    \label{app:qwen_object}
\end{figure*}

\begin{figure*}[t]
    \centering
    \includegraphics[width=\textwidth]{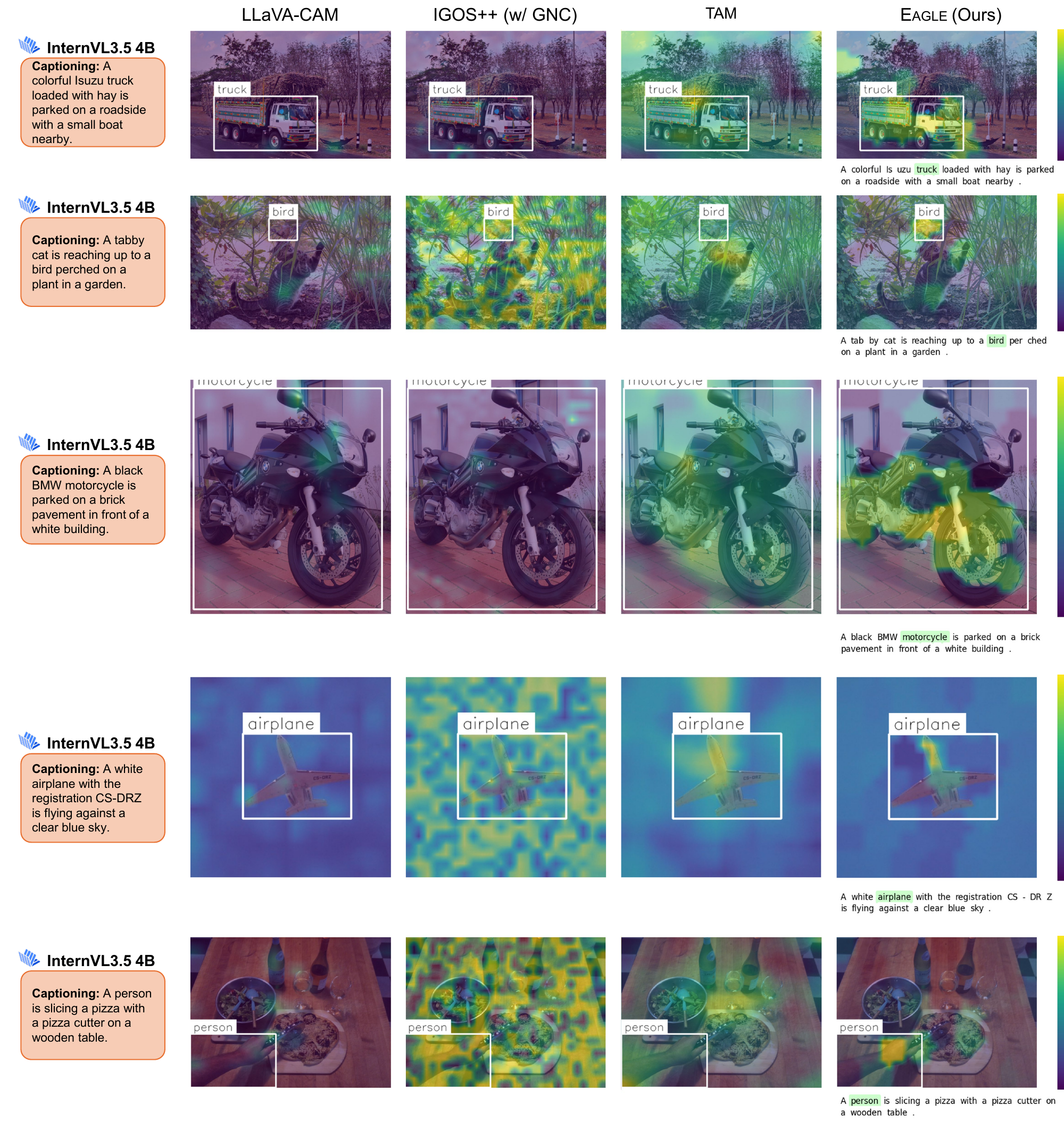}
    \caption{Word-level explanation results for \textbf{InternVL3.5} on the MS COCO dataset. Our method captures object-centric highlights with strong correspondence to caption tokens and bounding boxes.}
    \label{app:internvl_object}
\end{figure*}

Beyond sentence-level results, we further evaluate our method at the word-level with ground-truth bounding boxes. Fig.~\ref{app:llava_object}, Fig.~\ref{app:qwen_object}, and Fig.~\ref{app:internvl_object} illustrate that our method produces sparse yet highly accurate localization of queried objects such as ``boat,'' ``keyboard,'' or ``truck.'' By contrast, IGOS++ frequently covers overly broad regions, while LLaVA-CAM and TAM often fail to precisely localize objects. These comparisons highlight the advantage of our method in generating interpretable, object-centric attributions.

\subsection{Additional Hallucination Attribution Visualizations}

\begin{figure*}[t]
    \centering
    \includegraphics[width=\textwidth]{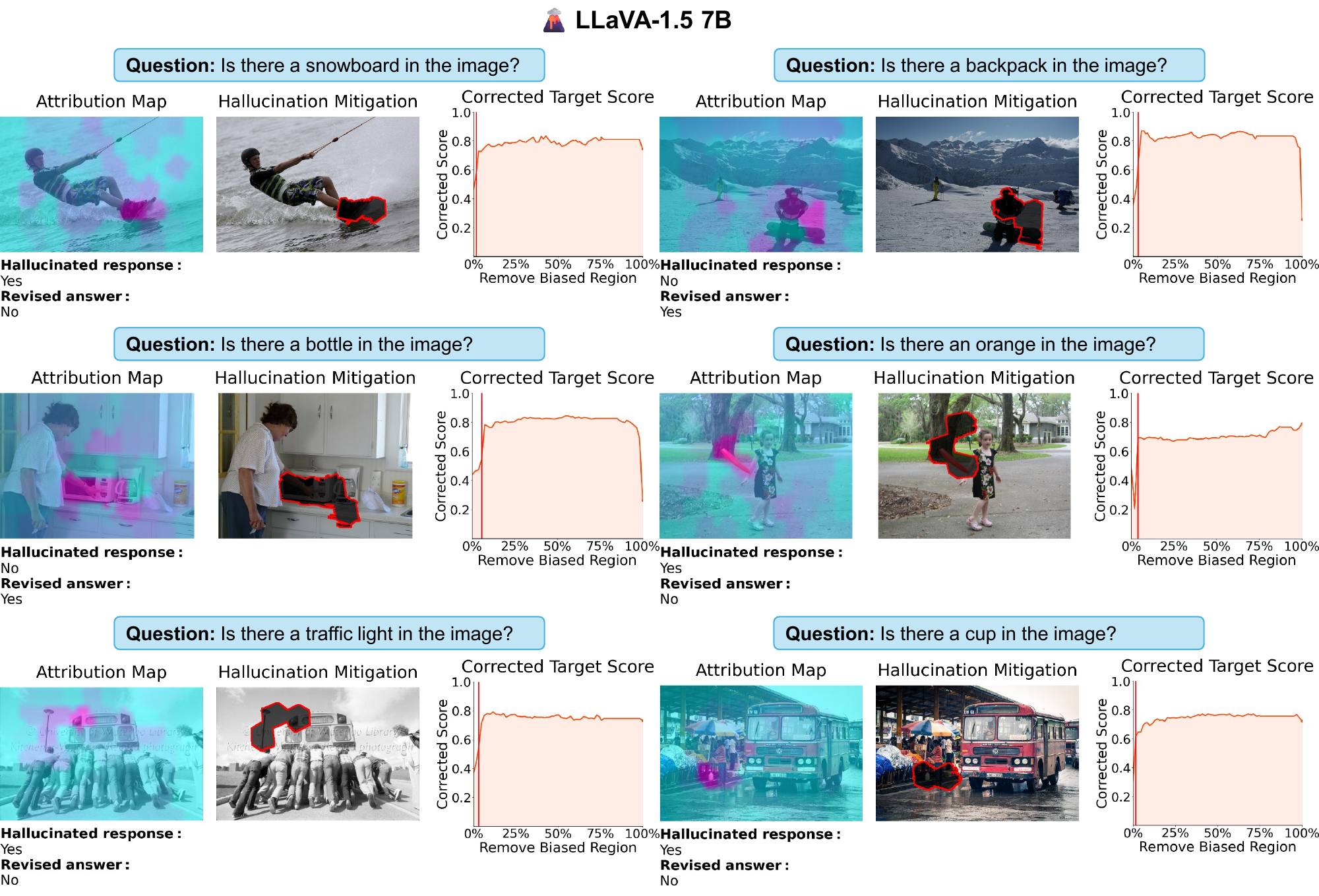}
    \caption{Hallucination attribution for \textbf{LLaVA-1.5} on the MS COCO dataset. Our method highlights the minimal hallucination-inducing regions across different queries, such as ``snowboard,'' ``traffic light,'' and ``cup.''}
    \label{app:hallucination_llava}
\end{figure*}

\begin{figure*}[t]
    \centering
    \includegraphics[width=\textwidth]{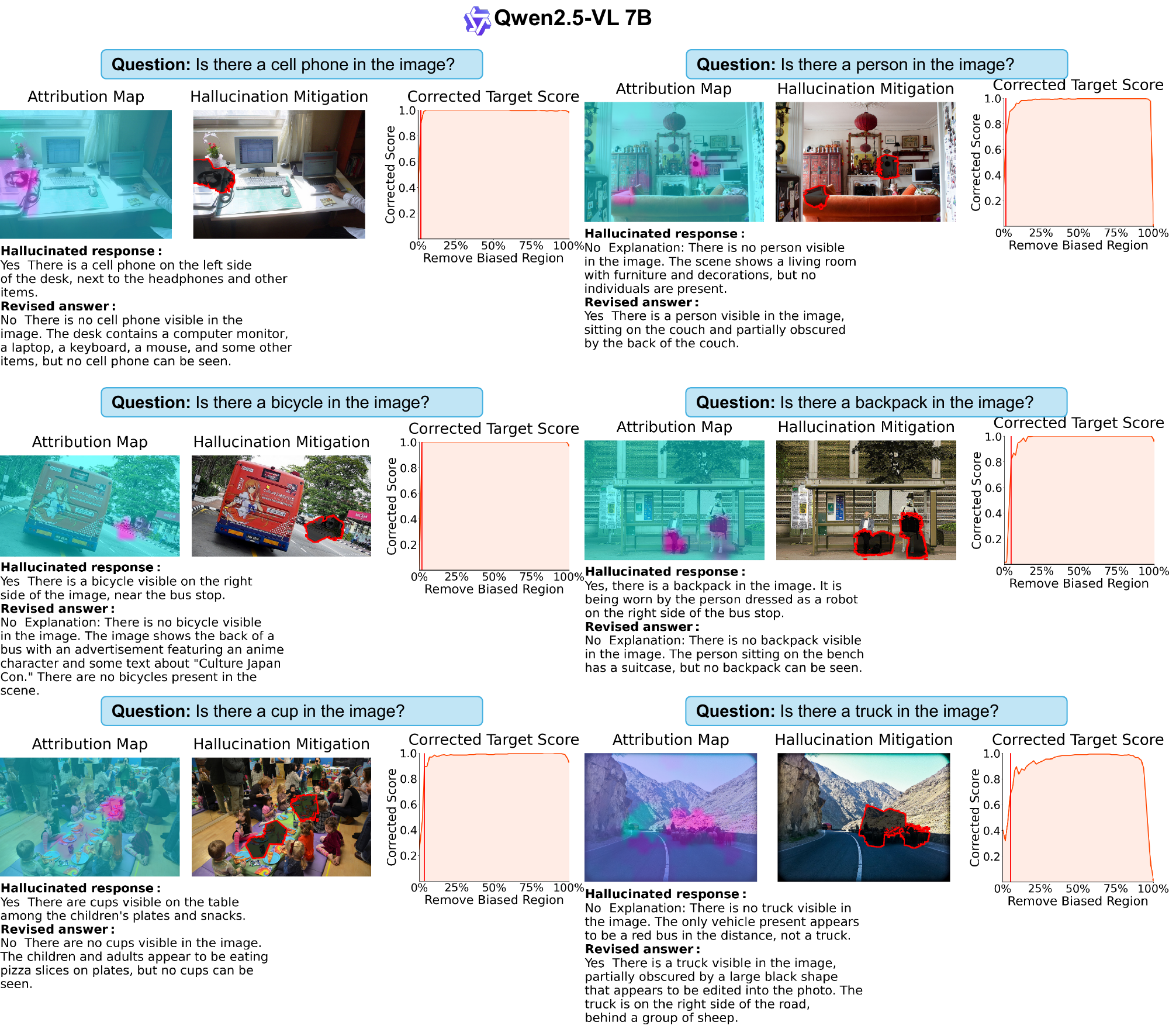}
    \caption{Hallucination attribution for \textbf{Qwen2.5-VL} on the MS COCO dataset. Our method isolates misleading cues leading to hallucinations in queries such as ``cell phone,'' ``bicycle,'' and ``truck.''}
    \label{app:hallucination_qwen}
\end{figure*}

\begin{figure*}[t]
    \centering
    \includegraphics[width=\textwidth]{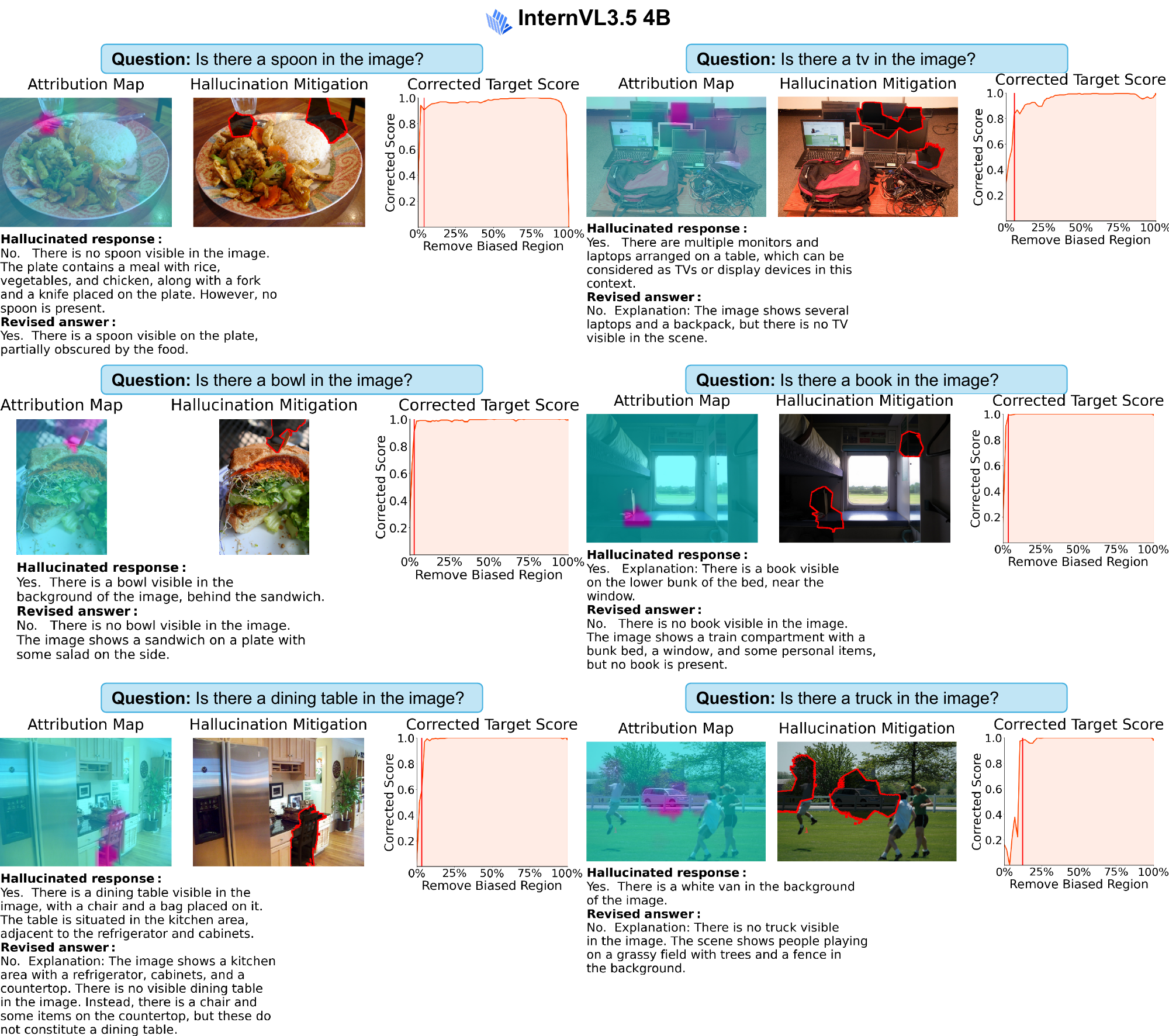}
    \caption{Hallucination attribution for \textbf{InternVL3.5} on the MS COCO dataset. Our method identifies hallucination-prone regions for queries such as ``spoon,'' ``tv,'' and ``dining table,'' especially in cases of overlapping or occluded objects.}
    \label{app:hallucination_internvl}
\end{figure*}

We also provide supplementary hallucination attribution results on MS COCO (Fig.~\ref{app:hallucination_llava}, Fig.~\ref{app:hallucination_qwen}, Fig.~\ref{app:hallucination_internvl}). Unlike the main paper, these figures focus exclusively on our method to illustrate how it identifies hallucination-prone regions across diverse queries.

For \textbf{LLaVA-1.5} (Fig.~\ref{app:hallucination_llava}), hallucinations typically arise from visually similar structures. For example, queries about a ``snowboard'' lead to confusions with surfboard-like regions, while small background cues induce false detections for ``traffic light'' or ``cup.'' Our attribution maps isolate these exact regions, providing interpretable evidence of failure modes.

For \textbf{Qwen2.5-VL} (Fig.~\ref{app:hallucination_qwen}), hallucinations are often caused by small or occluded objects. For instance, reflective regions resembling a phone screen mislead the model when asked about ``cell phones,'' while circular patterns in the background induce false positives for ``bicycle.'' Our approach sharply localizes these misleading cues, enhancing transparency.

Finally, for \textbf{InternVL3.5} (Fig.~\ref{app:hallucination_internvl}), hallucinations are triggered by overlapping or occluded objects. For example, confusion between a fork and a spoon is precisely localized, as are reflective regions falsely identified as ``TVs'' or cluttered areas misinterpreted as ``dining tables.'' These examples underscore the effectiveness of our method in diagnosing hallucination sources in a fine-grained and transparent manner.

\clearpage

\clearpage

\end{document}